\documentclass[11pt]{article}
\usepackage[margin=1in]{geometry}
\usepackage[T1]{fontenc}
\usepackage[utf8]{inputenc}
\usepackage{booktabs}
\usepackage{longtable}
\usepackage{array}
\usepackage{placeins}
\usepackage{graphicx}
\usepackage{amsmath}
\usepackage{xcolor}
\usepackage[numbers,sort&compress]{natbib}
\usepackage[hidelinks]{hyperref}
\newcommand{\rot}[1]{\rotatebox{90}{#1}}
\graphicspath{{figures/}}

\title{Evaluating Decision Models for Text Annotation in Computational Social Science}
\author{Hazem Ibrahim\thanks{Corresponding author: \texttt{hazem.ibrahim@nyu.edu}} \quad Yasir Zaki\\[4pt]
Computer Science, New York University Abu Dhabi, Abu Dhabi, UAE}
\date{}

\begin{document}
\maketitle

\begin{abstract}
Computational social science increasingly relies on large language models for text annotation, and the validity of published findings now rests on the labels generated by such models. Decision models, a new model class built for categorical question answering, answer typed questions with a choice, a probability distribution over the label set, and a confidence score rather than free text, at a small fraction of frontier inference prices. Whether their answers are accurate, and whether that stated confidence can be trusted on social science constructs, are unknown. Here, we mirror the evaluation of Ziems et al.\ (2024) on 18 computational social science classification tasks (7{,}977 items), comparing the first commercial decision model and two open-weight counterparts against 19 frontier and open-weight language models under the same zero-shot protocol, and extending the decision-model comparison to eleven open-weight systems released in the week after it. The decision model trails the per-task best LLM on 14 of 15 evaluation tasks, with a median deficit of 11.6 macro-F1 points, at a median 44 times lower measured cost. Its confidence is better calibrated than the verbalized confidence of 16 of the 19 LLMs, yet three frontier models show lower median calibration error (0.157 against 0.066). While items above 0.9 confidence are typically labeled accurately (median accuracy 0.815), on one task, empathy in peer-support dialogues, the model reports high confidence while performing near chance. Nonetheless, our results suggest that decision models are useful as a first step in the annotation pipeline: routing low-confidence items to an LLM matches or exceeds the LLM alone at a quarter to half of its cost.
\end{abstract}

\section{Introduction}

A growing share of empirical social science now rests on labels produced by large language models. Since Ziems et al.\ showed that zero-shot LLMs reach fair agreement with human annotators across a broad suite of computational social science (CSS) tasks \citep{ziems2024llms}, researchers have used LLMs to code stance, ideology, hate speech, and discourse structure in datasets far larger than any human team could annotate \citep{gilardi2023chatgpt, tornberg2023chatgpt}. The question is no longer whether a model can label social science text, but whether a researcher can know, for a given item, how much to trust the label \citep{pangakis2023automated}.

Frontier LLMs answer that question poorly for three reasons. First, chat-tuned models produce free text, so the label must be extracted from a generated answer, and a failed extraction produces a wrong label rather than an error. Second, the APIs serving frontier models rarely expose token-level probabilities (none of the 19 baseline APIs in our evaluation returned usable ones; Section~3.3), so calibration must be elicited verbally, and verbalized confidence is known to be poorly calibrated \citep{xiong2023can}. Third, frontier inference is priced for conversation, and labeling a corpus of several million items at frontier prices costs five to six figures, beyond most research budgets.

A new class of models addresses all three limitations by construction. Decision models accept a typed question with an enumerated label set and return a choice, a probability distribution over the full label set, and a scalar confidence score; they do not generate free text. The first commercially available decision model, TypeSafe's Jev, was released in September 2026 with input pricing of \$0.042 per million tokens; TypeSafe names its training objective reinforcement learning for calibrated decisions (RLCD) but has not published the method. An independent open-weight implementation of that objective, a 0.6B decision model built on Qwen3 and released four days later \citep{maio2026rlcd}, is available for local inference. The typed interface makes extraction errors impossible by construction, because the model can only return one of the declared options; the returned distribution makes calibration a directly measurable property of every labeling run rather than a separately elicited one; and the pricing changes what corpus sizes are feasible under constrained budgets. A decision model scores the declared options in a single forward pass rather than generating token by token, so small members of this class of model can run on hardware most research teams already own. Within days of Jev's release, open-weight reproductions spanning 0.15B to 9B parameters appeared; we evaluate eleven of them alongside the commercial model to test whether our findings describe one vendor's product or the class as a whole. RLCD here is distinct from reinforcement learning from contrastive distillation \citep{yang2023rlcd}, an unrelated alignment method that shares the acronym.

Whether these properties translate into accurate, trustworthy annotation on social science data is not obvious in either direction. CSS constructs are contested, socially situated, and often defined against codebooks that trained annotators disagree on. A confidence score calibrated on commercial decision workloads may not remain calibrated on implicit hate speech, on dialect features of Indian English, or on the discourse structure of Reddit arguments. If calibration holds, the annotation pipeline gains a component that labels at a small fraction of frontier prices and states, reliably, which items it should not label. If calibration fails, the confidence score would direct automation toward exactly the labels that most need human review.

Here, we answer that question by mirroring the evaluation of \citet{ziems2024llms} on 18 of their CSS classification tasks, the same released test splits, and the same zero-shot protocol, comparing the decision-model class (Jev, plus an open-weight 0.6B RLCD decision model and its no-training ablation on the same base) against 19 LLM baselines spanning the GPT-5.6, Claude, Gemini, Llama, Qwen, DeepSeek, Mistral, and GLM families. Beyond the accuracy comparison, we add three measurements that the 2023-era evaluation could not include: calibration (expected calibration error and reliability diagrams, computed from the returned distributions), coverage--accuracy routing curves (the accuracy of the model on the subset of items it is confident about, as a function of the confidence threshold), and measured cost per million labels. The design, hypotheses, task grid, and correction procedure were pre-registered (AsPredicted \#312{,}511), with three tasks used as a pilot and the remaining 15 evaluation tasks carrying all headline comparisons.

We find, first, that the commercial decision model trails the per-task best LLM on 14 of the 15 evaluation tasks, with a median deficit of 11.6 macro-F1 points, significant after false-discovery correction on 12 tasks; the confidence interval spans zero on ideological books, Indian English dialect, and character tropes. Second, its native calibration beats the verbalized confidence of 16 of the 19 LLM baselines but not that of the three frontier Claude models (median expected calibration error 0.157 against 0.066 for the best baseline). Third, items at or above 0.9 confidence reach a median accuracy of 0.815 at 0.376 coverage, but on the empathy task confidence carries no information (calibration error 0.538, accuracy in the confident subset 0.383 against a 0.371 base rate). Fourth, on the Indian English dialect task, where annotation models routinely perform worst, the decision model's calibration error of 0.063 is less than half its median across the other utterance-family tasks (0.157) and the lowest of the 22 models evaluated on that task. Finally, cascades that keep confident items on the decision model and escalate the rest to an LLM, match the partner LLM alone at roughly a quarter to half of its measured cost. Together, our results suggest that decision models like Jev are not yet substitutes for frontier LLM annotation, but can already serve as an inexpensive first stage of an annotation pipeline, provided their stated confidence is validated on each construct before it is trusted.

\section{Related Work}

\paragraph{LLMs as annotators in computational social science.}
Computational social science has long needed more labeled text than it can afford to annotate \citep{lazer2009computational}, and the field's response to each generation of language technology has been to test it as an annotator, from crowdsourcing \citep{snow2008cheap} to GPT-3-era few-shot labeling \citep{brown2020language, wang2021labeling, ding2023gpt3}. \citet{ziems2024llms} gave this practice its systematic foundation, evaluating 13 language models on 25 CSS benchmarks and concluding that zero-shot LLMs fail to outperform fine-tuned baselines on classification but reach agreement levels adequate for use inside human annotation teams; their released pipeline, prompts, and test splits define the foundation for this study. Parallel work reported that ChatGPT exceeded crowdworker accuracy on several political text tasks \citep{gilardi2023chatgpt, tornberg2023chatgpt} and that LLMs recover psychological constructs from text at scale \citep{rathje2024gpt}, with broader programs proposing LLMs as simulated respondents and research instruments \citep{argyle2023out, grossmann2023ai}. A validity literature followed on when LLM labels can substitute for human ones, covering error propagation into downstream estimates \citep{pangakis2023automated}, field-level risks of uncritical adoption \citep{ollion2024dangers}, and procedures for deciding when a model may replace an annotator \citep{calderon2025alternative}. Automated labels of this kind already carry large-scale measurement studies, from partisan-content audits of TikTok \citep{ibrahim2026tiktok} and stance detection on Twitter/X \citep{ibrahim2026stances} to LLM-based rewriting of news text \citep{kuo2025neutralizing}. At the same time, direct evaluations report task-dependent performance of LLMs against humans \citep{ibrahim2023perception} and measurable ideological slant and inconsistency in the models themselves \citep{aldahoul2025ideology}, which strengthens the case for validating model labels before they enter published estimates. This literature evaluates chat models through their generated text; decision models change the object of evaluation, because the model returns a distribution over the codebook rather than a string, and the validity question extends from accuracy to calibration.

\paragraph{Decision models.}
Decision models descend from the reinforcement-learning-from-feedback lineage that produced instruction-tuned assistants \citep{christiano2017deep, ouyang2022training, bai2022constitutional}, but they target a different objective, probabilities that match observed outcomes rather than text that raters prefer. TypeSafe's Jev \citep{typesafe2026jev} is the first such model offered commercially; TypeSafe has published the objective but not the reward, the training data, or the architecture. Rewarding calibrated confidence with a proper scoring rule has a peer-reviewed precedent in reasoning models \citep{damani2026beyond}. The open-weight 0.6B model we evaluate implements its own reading of the objective, a REINFORCE loop whose reward is the outcome minus the probability placed on the chosen option \citep{maio2026rlcd}; built on the Qwen model family \citep{qwen2024qwen25}, it permits local inference and, in this study, an ablation that isolates the contribution of the training from that of constrained decoding. The interface itself has older antecedents: zero-shot classifiers that score a label set supplied at inference time through natural language inference \citep{yin2019benchmarking, laurer2023building}, and task-specific reinforcement-learned estimators that return a probability rather than text \citep{nandakishor2025salesrlagent}. The training method, reinforcement learning for calibrated decisions, shares an acronym with reinforcement learning from contrastive distillation \citep{yang2023rlcd}, a preference-data generation method for alignment; the two are unrelated, and we use the full phrase where confusion is possible. Within days of Jev's release, an ecosystem of open-weight reproductions appeared, spanning frozen-model logit readouts \citep{lee2026semif}, LoRA adapters on open decoders \citep{bespoke2026nimble,palmer2026kev}, purpose-trained small decoders \citep{mapika2026decider}, encoder-based implementations \citep{kishor2026laya,panisa2026von,kumar2026verdict}, and wrappers around earlier zero-shot NLI classifiers \citep{wadhwa2026opendecision,laurer2023building}, tracked by a community leaderboard \citep{jevbench2026} concurrent with this study; we evaluate eleven of these systems on our grid in Appendix~\ref{app:exploratory}. We are not aware of any published evaluation of decision models on social science annotation.

\paragraph{Selective prediction and calibration.}
Routing items by model confidence is selective prediction, a problem with a fifty-year history beginning with the reject option in pattern recognition \citep{chow1970optimum} and formalized for deep networks as selective classification \citep{geifman2017selective}. Its usefulness depends entirely on calibration, conventionally measured by expected calibration error \citep{naeini2015obtaining}, and modern neural networks are known to be miscalibrated by default \citep{guo2017calibration}. For language models the evidence is mixed: larger models show partial knowledge of what they know when probabilities are read from logits \citep{kadavath2022language} and can be taught to express uncertainty in words \citep{lin2022teaching}, but many commercial serving APIs no longer expose token probabilities, and verbalized confidence elicited from chat models is systematically overconfident \citep{xiong2023can, tian2023just}.

\paragraph{Annotation economics and model routing.}
Routing each item to the cheapest system that can handle it is an established design for reducing annotation cost: model cascades escalate items by predicted difficulty or confidence \citep{chen2023frugalgpt, ding2024hybrid}, and related work studies when to defer to a human expert \citep{mozannar2020consistent} and how to set confidence thresholds in human--LLM verification pipelines \citep{wang2024human}. Every design in this family takes a confidence signal as input, so its guarantees depend on that signal being calibrated. Decision models supply the signal natively, because the distribution the router needs is the model's standard output, at a small fraction of frontier chat prices. Here, we test whether that distribution remains calibrated on CSS constructs.

\section{Methods}

\subsection{Mirror design}
We reuse the evaluation suite of \citet{ziems2024llms}: 18 closed-choice classification tasks, the same released test splits (class-stratified samples of at most 500 instances per task; 7{,}977 items in total), the same gold labels, and their prompting guidelines. Their published zero-shot results for FLAN-T5, GPT-3-series, and GPT-4 models, and their fine-tuned RoBERTa-large baselines, serve as historical comparison columns; we do not rerun their models. The task grid spans ten utterance-level tasks (dialect features \citep{demszky2021dialect}, emotion \citep{saravia2018carer}, figurative language \citep{chakrabarty2022flute}, implicit hate \citep{elsherief2021latent}, humor \citep{weller2019humor}, ideology \citep{iyyer2014political}, misinformation \citep{gabriel2022misinfo}, persuasion \citep{althoff2014favor}, stance \citep{mohammad2016semeval}, and semantic change \citep{loureiro2022tempowic}), six conversation-level tasks (discourse acts \citep{zhang2017coarse}, empathy \citep{sharma2020empathy}, persuasion strategies \citep{yang2019persuasive}, power \citep{danescu2012power}, toxicity prediction \citep{zhang2018awry}, and politeness \citep{danescu2013politeness}, several drawn through ConvoKit \citep{chang2020convokit}), and two document-level tasks, media ideology \citep{baly2020ideology} and character tropes \citep{bamman2013personas, chu2018personas}. Their five generation tasks are excluded by construction, because decision models do not generate free text, and two classification tasks whose released format is not closed-choice, hippocorpus (variable-length sentence-set extraction) and wikievents (free-text JSON slot filling), are excluded on the same ground. We treat this as a division of labor rather than a deficiency of the comparison: decision models are candidates for the labeling half of the CSS pipeline, and the explanatory half remains with generative models.

\subsection{Models}
We evaluate three decision-class models: Jev 1.13 (TypeSafe, via the OpenRouter decisions endpoint; the resolved model version and serving provider are recorded with every call), an open-weight 0.6B decision model trained under an independent implementation of the RLCD objective on Qwen3-0.6B-Base (local inference on university HPC), and a no-training ablation that applies the identical letter-token constrained-decoding pattern to the same vanilla base model, isolating the effect of RLCD training. Against these we run 19 LLM baselines via the OpenRouter, spanning the GPT-5.6, Claude, Gemini, Llama, Qwen, DeepSeek, Mistral, GLM, Kimi, and Gemma families; the full roster with resolved versions appears in the appendix reproducibility table. OpenAI and Gemini baselines run at reasoning effort ``low''; Kimi K2.6 runs with reasoning disabled and GLM 5.3 at reasoning effort ``low'', after both produced widespread empty completions from reasoning-token exhaustion in a first pass; all other LLM baselines run at temperature zero. LLM baselines receive Ziems et al.'s released prompts verbatim, plus one appended line eliciting a verbalized confidence from 0 to 100; a robustness check on the pilot tasks shows the appended line does not drive the accuracy comparisons, changing accuracy by a median of $+0.2$ points across the 48 pilot-task cells (median absolute change 1.7 points; Table~\ref{tab:confline}). Decision models receive the same instruction text and the same option descriptions, restated in the decisions API format as a typed choice question whose criteria are keyed by the gold label names. The mapping from each task's multiple-choice options to gold labels is fixed in code and was checked by hand against the gold label set. After the analysis plan was filed, eleven open-weight decision systems released in the week after Jev's launch (Laya, Von, Verdict, OpenDecision, decider-0.8b, decider-2b, SemIf, Bespoke-Nimble-9B, and Kev-0.8B, 4B, and 9B, spanning 0.15B to 9B parameters) were run on the same grid under each system's own released inference code; their construction and results are reported separately in Appendix~\ref{app:exploratory}.

\subsection{Metrics}
We report four measurement outcomes per task and model. First, accuracy and macro-F1 against gold labels, the metric of \citet{ziems2024llms}, computed on identical test items. Second, calibration: expected calibration error with 15 equal-width bins, computed on the decision models' returned confidence and on the probability of the chosen option, with reliability diagrams per task; LLM baselines, whose serving APIs return no usable token probabilities, are evaluated on verbalized confidence elicited in the same call and labeled as such. Third, coverage--accuracy routing curves: for each confidence threshold $t \in \{0.5, 0.6, 0.7, 0.8, 0.9, 0.95\}$, the fraction of items at or above threshold (coverage) and the accuracy within that fraction. Fourth, measured cost in dollars per task and extrapolated per million labels; inference pricing varies across providers and over time, so every cost figure reports the price measured at the time of the study (September 2026), taken from the per-call billing records. Calibration throughout is measured against the single adjudicated gold label per item, so our results speak to calibration against benchmark gold standards rather than to the underlying constructs, on which trained annotators themselves disagree; the one grid task with released per-annotator scores is tested against them directly in Appendix~\ref{app:exploratory}.

\subsection{Analysis plan and inference}
Our pre-registration (AsPredicted \#312{,}511) fixes the full grid (tasks $\times$ models $\times$ metrics), one characterization question, three hypotheses, and the confidence-cascade analysis, including the routing rule, both partner-selection rules (best frontier LLM and best LLM under \$0.50 per 1{,}000 items), and the evaluated threshold grid. The characterization question asks on which tasks Jev beats the best LLM baseline, estimated as per-task $\Delta$F1 with paired-bootstrap 95 percent confidence intervals and related to five task features, without a directional prediction. The three hypotheses are that (1) Jev's median expected calibration error is lower than the verbalized-confidence calibration error of every LLM baseline; (2) that on the median evaluation task, items with Jev confidence of at least 0.9 reach accuracy of at least 0.85 at coverage of at least 0.15; and (3) that Jev's calibration error on the Indian English dialect task exceeds its median over the other utterance-family evaluation tasks. The three pilot tasks are stance, implicit hate, and discourse acts; the remaining 15 evaluation tasks carry every headline comparison. Model-to-model F1 differences are tested with a paired bootstrap over items (10{,}000 resamples, fixed seed) per task, with Benjamini--Hochberg false discovery rate control at $q = 0.05$ across the full task-by-comparison grid \citep{benjamini1995controlling}. We note three departures from the pre-registration. First, the character-tropes cell is unrunnable for the two 0.6B decision models, whose architecture decodes over 26 letter tokens, and is reported as N/A. Second, the Kimi and GLM inference configurations were adjusted after reasoning-token exhaustion (Section~3.2). Lastly, per-call timestamps were not recorded, so throughput is not reported. Exploratory analyses appear in Appendix~\ref{app:exploratory}.

\section{Results}

\begin{table*}[t]
\centering
\caption{Macro-F1 ($\times$100) on the 18 CSS classification tasks. The Rand (random baseline), FT (fine-tuned RoBERTa-large), and GPT-4 columns reproduce the published 2023 values of \citet{ziems2024llms}; the remaining columns are our 2026 zero-shot runs on identical test items. Bold marks the best 2026 model per task; $^{d}$ marks the three pilot tasks, which are excluded from the headline comparisons; -- marks the character-tropes cells unrunnable for the 0.6B decision models and the fine-tuned value Ziems et al.\ did not release. Paired-bootstrap confidence intervals and BH-FDR-adjusted $p$-values for every model-to-model comparison appear in Appendix~\ref{app:registered}, Table~\ref{tab:full}.}
\label{tab:main}
\resizebox{\textwidth}{!}{\begin{tabular}{lrrrrrrrrrrrrrrrrrrrrrrrrr}
\toprule
 & \multicolumn{3}{c}{Ziems et al.\ (2023 runs)} & \multicolumn{22}{c}{2026 runs (ours)} \\
\cmidrule(lr){2-4}\cmidrule(lr){5-26}
Task & Rand & FT & GPT-4 & \rot{Jev} & \rot{Qwen3-0.6B-RLCD} & \rot{Qwen3-0.6B base} & \rot{Fable 5.1} & \rot{Haiku 4.5} & \rot{Opus 5} & \rot{Sonnet 5} & \rot{DS V3.2} & \rot{Gem 3.1P} & \rot{Gem FL} & \rot{Gem 3.8F} & \rot{Gemma 31B} & \rot{L4 Mav} & \rot{L4 Scout} & \rot{Mistral M} & \rot{Kimi 2.6} & \rot{Luna} & \rot{Sol} & \rot{Terra} & \rot{OSS 120B} & \rot{Qwen 235B} & \rot{GLM 5.3} \\
\midrule
\multicolumn{26}{l}{\itshape Utterance-level tasks} \\
Dialect & 3.3 & 3.0 & 23.2 & \textbf{64.0} & 5.7 & 2.7 & 52.1 & 39.6 & 58.2 & 56.8 & 46.4 & 59.7 & 50.4 & 59.8 & 54.8 & 33.8 & 36.6 & 38.6 & 44.0 & 56.1 & 55.8 & 53.7 & 51.3 & 42.4 & 53.9 \\
Emotion & 16.7 & 71.6 & 50.6 & 48.4 & 49.7 & 28.1 & 51.2 & 45.5 & \textbf{61.9} & 47.6 & 39.3 & 48.2 & 49.6 & 56.5 & 44.9 & 45.1 & 40.9 & 44.8 & 40.6 & 48.7 & 53.5 & 52.6 & 47.3 & 46.3 & 42.6 \\
Figurative & 25.0 & 99.2 & 17.5 & 86.3 & 30.1 & 27.4 & 91.5 & 78.5 & 92.0 & 88.8 & 81.1 & 90.8 & 90.8 & \textbf{92.5} & 88.2 & 79.7 & 66.0 & 82.4 & 84.7 & 86.5 & 89.7 & 89.0 & 87.5 & 81.5 & 88.6 \\
Humor & 49.5 & 73.1 & 61.3 & 56.3 & 44.2 & 33.1 & 64.1 & 61.7 & 60.4 & \textbf{69.3} & 54.9 & 63.3 & 63.6 & 67.7 & 61.9 & 57.4 & 57.2 & 53.8 & 63.2 & 62.0 & 64.6 & 64.0 & 43.3 & 60.4 & 58.1 \\
Ideology (utt.) & 33.3 & 64.8 & 60.0 & 63.3 & 19.6 & 19.7 & 61.2 & 60.7 & 60.4 & 59.7 & 63.6 & 64.3 & \textbf{65.3} & 63.2 & 64.6 & 61.1 & 48.6 & 61.7 & 58.3 & 61.4 & 65.0 & 61.8 & 60.6 & 62.8 & 62.9 \\
Impl. Hate$^{d}$ & 16.7 & 62.5 & 3.7 & 43.9 & 28.0 & 17.6 & 55.9 & 32.5 & \textbf{59.7} & 50.1 & 33.1 & 55.0 & 49.5 & 57.1 & 47.9 & 35.1 & 26.7 & 17.8 & 42.1 & 50.2 & 54.7 & 52.6 & 35.6 & 46.2 & 48.0 \\
Misinfo & 50.0 & 81.6 & 26.9 & 79.6 & 60.2 & 34.2 & 82.3 & 82.0 & 83.0 & 80.8 & 79.8 & 83.4 & \textbf{83.9} & 82.1 & 80.4 & 78.7 & 81.0 & 80.7 & 83.6 & 81.5 & 81.5 & 82.0 & 80.8 & 81.0 & 83.0 \\
Persuasion (utt.) & 14.3 & 52.0 & 56.4 & 60.7 & 39.2 & 6.3 & 68.2 & 52.1 & 68.8 & 61.8 & 47.2 & 66.7 & 54.8 & \textbf{72.4} & 53.9 & 48.4 & 48.2 & 51.4 & 58.1 & 57.7 & 66.9 & 64.2 & 55.1 & 49.4 & 60.6 \\
Sem. Change & 50.0 & 62.3 & 21.2 & 68.3 & 44.9 & 41.3 & 82.2 & 59.8 & 76.0 & 82.2 & 63.6 & 84.7 & 73.2 & \textbf{84.9} & 70.5 & 65.3 & 60.0 & 71.2 & 70.5 & 65.0 & 79.6 & 77.3 & 71.2 & 62.7 & 76.8 \\
Stance$^{d}$ & 33.3 & 36.1 & 76.0 & 73.4 & 48.2 & 16.7 & 68.8 & 70.5 & 65.5 & 70.0 & 62.9 & 74.2 & 67.6 & \textbf{75.8} & 64.2 & 56.1 & 61.5 & 59.0 & 60.4 & 64.5 & 73.0 & 61.7 & 74.3 & 62.9 & 56.5 \\
\midrule
\multicolumn{26}{l}{\itshape Conversation-level tasks} \\
Discourse$^{d}$ & 14.3 & 49.6 & 16.7 & 59.5 & 27.5 & 8.2 & 69.0 & 57.3 & 69.7 & 64.2 & 56.1 & 70.4 & 64.1 & \textbf{71.1} & 60.0 & 50.5 & 45.7 & 61.0 & 66.7 & 61.3 & 67.6 & 67.0 & 55.7 & 55.4 & 63.9 \\
Empathy & 33.3 & 71.6 & 6.4 & 26.9 & 30.7 & 16.9 & 29.9 & 26.3 & 30.4 & 27.7 & 22.8 & \textbf{31.4} & 25.0 & 27.1 & 25.2 & 20.6 & 30.6 & 22.6 & 23.1 & 27.3 & 27.3 & 24.3 & 24.0 & 22.9 & 30.2 \\
Persuasion (conv.) & 50.0 & 33.3 & 28.6 & 56.1 & 44.6 & 33.3 & 73.7 & 65.9 & 72.7 & 65.2 & 51.6 & \textbf{75.0} & 63.2 & 72.5 & 61.1 & 48.9 & 50.1 & 56.6 & 59.3 & 56.3 & 68.2 & 63.4 & 40.6 & 60.7 & 64.0 \\
Politeness & 33.3 & 75.8 & 59.7 & 57.3 & 27.0 & 22.6 & 65.3 & 61.1 & 63.9 & 66.8 & 66.7 & \textbf{67.8} & 62.5 & 66.9 & 64.0 & 58.1 & 44.7 & 45.1 & 61.7 & 59.7 & 58.1 & 62.9 & 56.9 & 50.9 & 53.5 \\
Power & 49.5 & 72.7 & 42.0 & 58.1 & 38.8 & 43.7 & 69.4 & 64.5 & \textbf{71.1} & 65.6 & 55.7 & 63.4 & 66.7 & 66.3 & 60.4 & 59.7 & 60.9 & 54.6 & 67.8 & 56.6 & 63.1 & 64.2 & 52.5 & 61.2 & 69.7 \\
Toxicity & 50.0 & 64.6 & 55.4 & 50.2 & 33.3 & 41.9 & 59.0 & 59.9 & 60.6 & 53.8 & 41.7 & 57.9 & 60.5 & 61.7 & 58.9 & 59.8 & 56.7 & 56.0 & 57.1 & 58.2 & \textbf{63.0} & 62.4 & 51.9 & 49.0 & 56.4 \\
\midrule
\multicolumn{26}{l}{\itshape Document-level tasks} \\
Ideology (doc.) & 33.3 & 85.1 & 51.5 & 65.4 & 22.6 & 17.4 & 76.8 & 59.3 & 79.3 & 68.6 & 64.3 & 79.7 & 69.5 & \textbf{82.6} & 64.6 & 56.4 & 46.0 & 68.2 & 71.3 & 67.1 & 76.7 & 75.1 & 57.8 & 63.0 & 64.9 \\
Tropes & 36.9 & -- & 44.9 & 19.0 & -- & -- & 20.4 & 3.0 & \textbf{26.2} & 13.5 & 13.4 & 24.7 & 19.5 & 24.3 & 14.5 & 9.2 & 10.5 & 15.0 & 12.6 & 22.2 & 25.5 & 21.3 & 11.0 & 14.9 & 23.3 \\
\bottomrule
\end{tabular}
}
\end{table*}

\subsection{Accuracy against gold labels}

We begin by examining annotation accuracy, the part of the evaluation that mirrors \citet{ziems2024llms} directly. Table~\ref{tab:main} reports macro-F1 for the three decision-class models and the 19 LLM baselines on all 18 tasks, alongside the published 2023 values for the random baseline, the fine-tuned RoBERTa-large models, and GPT-4. We compare Jev against the per-task best LLM with paired-bootstrap confidence intervals, and we test all 342 task-by-comparison cells with BH-FDR correction at $q = 0.05$; 194 cells are significant after correction.

\begin{figure*}[tbp]
\centering
\begin{minipage}[t]{0.485\textwidth}
\centering
\includegraphics[width=\linewidth]{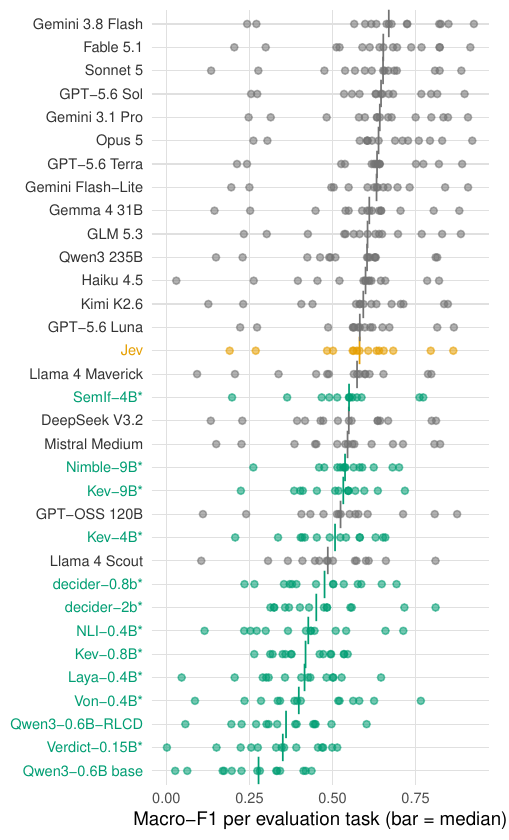}
\caption{Macro-F1 on the evaluation tasks, by model (dots: tasks; bar: median). Orange marks Jev, green the other decision models, grey the 19 LLM baselines. Rows marked * are open decision systems added after the analysis plan was filed (Appendix~\ref{app:exploratory}). Decision models other than Jev are scored on 14 evaluation tasks (13 for SemIf) because the 114-option tropes task exceeds their answer slots.}
\label{fig:accuracy}
\end{minipage}\hfill
\begin{minipage}[t]{0.485\textwidth}
\centering
\includegraphics[width=\linewidth]{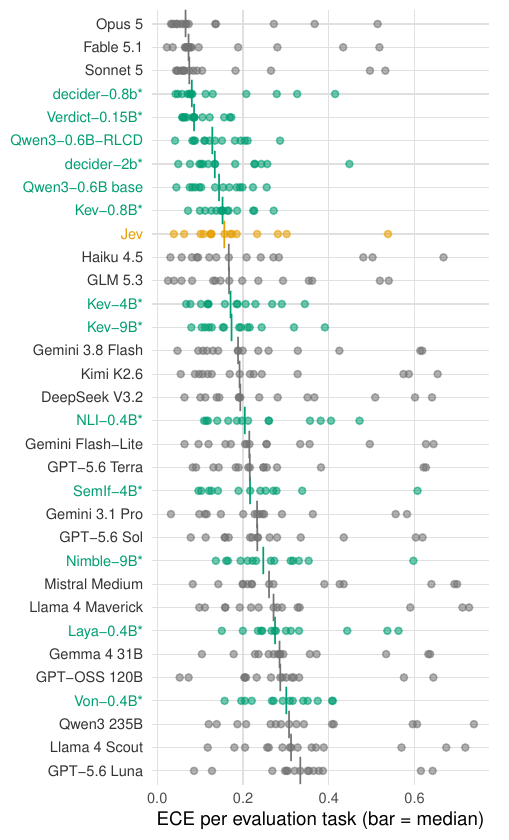}
\caption{Expected calibration error on the evaluation tasks, by model (dots: tasks; bar: median). Orange marks Jev, green the other decision models, grey the 19 LLM baselines evaluated on verbalized confidence. Rows marked * are open decision systems added after the analysis plan was filed (Appendix~\ref{app:exploratory}). Decision models other than Jev are scored on 14 evaluation tasks (13 for SemIf) because the 114-option tropes task exceeds their answer slots.}
\label{fig:calibration}
\end{minipage}
\end{figure*}

Across the 15 evaluation tasks, Jev reaches a median accuracy of 0.610 and a median macro-F1 of 0.581, a median deficit of 11.6 macro-F1 points to the per-task best LLM, and the deficit is significant after correction on 12 of the 15 tasks. The per-task best is selected on the same test items, an oracle comparator that favors the LLM side of every comparison. On two further tasks Jev is statistically comparable to the best LLM: ideological books ($\Delta$F1 $= -2.1$ points, CI $[-5.9, +1.8]$) and character tropes ($-7.1$ points, CI $[-14.4, +0.3]$). On the remaining task, Indian English dialect, Jev posts the best point estimate in the grid ($+4.2$ points over the best LLM), though the confidence interval spans zero here as well (CI $[-3.9, +11.5]$). A selection-aware bootstrap that re-selects the best LLM inside each resample (Appendix~\ref{app:exploratory}) leaves the ideological-books and dialect intervals spanning zero but moves the character-tropes interval below it ($[-15.5, -1.4]$ points), so that exception is sensitive to the selection step. The historical columns place these numbers in context: the best 2026 zero-shot model exceeds Ziems et al.'s GPT-4 on every task except stance and character tropes, yet still trails their fine-tuned RoBERTa-large on 9 of the 17 tasks with a released fine-tuned value, so the 2023 conclusion that zero-shot models do not replace supervised classifiers survives three years of model progress. Figure~\ref{fig:accuracy} places this median in the full grid: it exceeds the medians of five LLMs (both Llama 4 models, DeepSeek V3.2, Mistral Medium, and GPT-OSS 120B) and of every other decision model, the strongest of which (SemIf-4B, 0.550) exceeds four.

Five task characteristics, the number of answer options, the balance of the label distribution, the median context length, the task family, and the difficulty of the task for Ziems et al.'s fine-tuned baseline, explain little of the variation in $\Delta$F1. None of the four rank correlations is significant at $n = 15$ (number of options $\rho = +0.41$, $p = 0.13$; label balance $\rho = -0.36$, $p = 0.19$; context length $\rho = -0.44$, $p = 0.10$; fine-tuned F1 $\rho = +0.01$, $p = 0.97$), and the family medians are close (utterance $-8.9$, conversation $-12.8$, document $-12.2$ points). Within the decision-model class, RLCD training raises the 0.6B model's median macro-F1 from 0.278 to 0.361 for the identical base model under identical constrained decoding, though both remain far below Jev's 0.581.

\begin{table}[tbp]
\centering
\caption{Calibration and routing summary over the 15 evaluation tasks: median expected calibration error (15 equal-width bins), median Brier score, median area under the coverage--accuracy curve, and accuracy and coverage at the 0.9 confidence threshold on the median task. Decision-class models are evaluated on their returned confidence; LLMs on verbalized confidence elicited in the same call. -- marks the base model's empty confident subset. $^{\dagger}$ marks open decision systems added after the analysis plan was filed (Appendix~\ref{app:exploratory}); like the two 0.6B models they cover 14 evaluation tasks (SemIf 13).}
\label{tab:calibration}
\begin{tabular}{lrrrrr}
\toprule
Model & Med.\ ECE & Med.\ Brier & Med.\ AUC & Acc@0.9 & Cov@0.9 \\
\midrule
Opus 5 & 0.066 & 0.201 & 0.783 & 0.862 & 0.259 \\
Fable 5.1 & 0.073 & 0.204 & 0.796 & 0.953 & 0.192 \\
Sonnet 5 & 0.075 & 0.214 & 0.766 & 0.942 & 0.095 \\
decider-0.8b$^{\dagger}$ & 0.081 & 0.619 & 0.590 & 0.804 & 0.031 \\
Verdict-0.15B$^{\dagger}$ & 0.086 & 0.668 & 0.486 & 0.833 & 0.000 \\
Qwen3-0.6B-RLCD & 0.128 & 0.662 & 0.505 & 0.625 & 0.000 \\
decider-2b$^{\dagger}$ & 0.134 & 0.637 & 0.599 & 0.780 & 0.034 \\
Qwen3-0.6B base & 0.144 & 0.688 & 0.445 & -- & 0.000 \\
Kev-0.8B$^{\dagger}$ & 0.153 & 0.628 & 0.550 & 0.882 & 0.000 \\
Jev & 0.157 & 0.539 & 0.775 & 0.815 & 0.376 \\
Haiku 4.5 & 0.167 & 0.249 & 0.744 & 0.852 & 0.191 \\
GLM 5.3 & 0.168 & 0.252 & 0.768 & 0.848 & 0.376 \\
Kev-4B$^{\dagger}$ & 0.171 & 0.616 & 0.668 & 0.891 & 0.031 \\
Kev-9B$^{\dagger}$ & 0.173 & 0.574 & 0.675 & 0.986 & 0.010 \\
Gem 3.8F & 0.189 & 0.237 & 0.795 & 0.797 & 0.742 \\
Kimi 2.6 & 0.192 & 0.257 & 0.741 & 0.901 & 0.159 \\
DS V3.2 & 0.194 & 0.272 & 0.680 & 0.679 & 0.167 \\
NLI-0.4B$^{\dagger}$ & 0.204 & 0.688 & 0.489 & 0.962 & 0.000 \\
Gem FL & 0.215 & 0.267 & 0.712 & 0.736 & 0.488 \\
Terra & 0.216 & 0.262 & 0.761 & 0.735 & 0.628 \\
SemIf-4B$^{\dagger}$ & 0.217 & 0.572 & 0.679 & 0.682 & 0.264 \\
Gem 3.1P & 0.233 & 0.260 & 0.801 & 0.737 & 0.710 \\
Sol & 0.233 & 0.267 & 0.759 & 0.738 & 0.763 \\
Nimble-9B$^{\dagger}$ & 0.248 & 0.663 & 0.662 & 0.690 & 0.524 \\
Mistral M & 0.261 & 0.287 & 0.696 & 0.621 & 0.700 \\
L4 Mav & 0.271 & 0.304 & 0.656 & 0.767 & 0.287 \\
Laya-0.4B$^{\dagger}$ & 0.275 & 0.693 & 0.499 & 0.555 & 0.000 \\
Gemma 31B & 0.285 & 0.301 & 0.716 & 0.660 & 0.815 \\
OSS 120B & 0.287 & 0.301 & 0.654 & 0.683 & 0.371 \\
Von-0.4B$^{\dagger}$ & 0.301 & 0.582 & 0.539 & 0.726 & 0.000 \\
Qwen 235B & 0.307 & 0.332 & 0.669 & 0.628 & 0.853 \\
L4 Scout & 0.312 & 0.333 & 0.619 & 0.653 & 0.526 \\
Luna & 0.333 & 0.331 & 0.708 & 0.637 & 0.862 \\
\bottomrule
\end{tabular}

\end{table}

\subsection{Calibration of stated confidence}

We next examine whether the confidence a model states can be believed, a measurement the 2023 evaluation could not include. For the decision-class models we compute expected calibration error directly from the returned distributions; for the LLM baselines we compute it from verbalized confidence. Table~\ref{tab:calibration} and Figure~\ref{fig:calibration} summarize the grid.

Our calibration hypothesis, that the decision model's calibration error would be lower than that of every baseline, holds against 16 of the 19 baselines but fails against the three frontier Claude models, and is therefore unsupported as stated. Jev's median expected calibration error over the evaluation tasks is 0.157 (95 percent task-level bootstrap CI $[0.108, 0.233]$), lower than the verbalized-confidence error of 16 of the 19 LLM baselines but higher than that of the three frontier Claude models (Opus 5 at 0.066, CI $[0.046, 0.137]$; Fable 5.1 at 0.073; Sonnet 5 at 0.075). The median intervals overlap, reflecting a median over 15 tasks of at most 500 items each; per-task intervals appear in Appendix~\ref{app:exploratory}. These comparisons use 15 equal-width bins, and are conditional on the model returning a usable confidence: items whose verbalized confidence is absent or negative are kept for accuracy but excluded from calibration and routing (Table~\ref{tab:exclusions}; DeepSeek V3.2 declines to estimate on 55 percent of items), whereas the decision models return a distribution on every call. Under equal-mass binning the frontier margin narrows and Sonnet 5 falls slightly behind Jev, while Opus 5 stays ahead under every binning (Appendix~\ref{app:exploratory}). The hypothesis required Jev to beat every baseline, and the models that beat it do so on verbalized confidence, a signal the calibration literature characterizes as systematically overconfident \citep{xiong2023can, tian2023just}; on this suite it nonetheless shows a lower median calibration error than the decision model's native distribution. Jev's miscalibration is concentrated rather than uniform: the empathy task contributes a calibration error of 0.538, and removing it lowers the median only to 0.142, still above the three frontier Claude point estimates. The 0.6B RLCD model reports a lower median calibration error than Jev (0.128), but its median Brier score of 0.662 and coverage--accuracy area of 0.505 suggest a confidence signal with almost no ranking value.

\begin{table}[tp]
\centering
\caption{Measured cost per 1{,}000 items and total cost for the full 18-task grid (7{,}977 items per model). Local models run on university hardware and carry no metered cost; their prices are hosting-dependent rather than zero (Section~5). $^{\dagger}$ marks open decision systems added after the analysis plan was filed (Appendix~\ref{app:exploratory}); like the two 0.6B models they cover 17 of the 18 tasks (SemIf 16).}
\label{tab:cost}
\begin{tabular}{lrr}
\toprule
Model & \$/1{,}000 items & Total grid cost (USD) \\
\midrule
Jev & 0.027 & 0.21 \\
Qwen3-0.6B-RLCD (local) & 0.000 & 0.00 \\
Qwen3-0.6B base (local) & 0.000 & 0.00 \\
Verdict-0.15B$^{\dagger}$ (local) & 0.000 & 0.00 \\
Laya-0.4B$^{\dagger}$ (local) & 0.000 & 0.00 \\
NLI-0.4B$^{\dagger}$ (local) & 0.000 & 0.00 \\
Von-0.4B$^{\dagger}$ (local) & 0.000 & 0.00 \\
decider-0.8b$^{\dagger}$ (local) & 0.000 & 0.00 \\
Kev-0.8B$^{\dagger}$ (local) & 0.000 & 0.00 \\
decider-2b$^{\dagger}$ (local) & 0.000 & 0.00 \\
SemIf-4B$^{\dagger}$ (local) & 0.000 & 0.00 \\
Kev-4B$^{\dagger}$ (local) & 0.000 & 0.00 \\
Nimble-9B$^{\dagger}$ (local) & 0.000 & 0.00 \\
Kev-9B$^{\dagger}$ (local) & 0.000 & 0.00 \\
Fable 5.1 & 7.897 & 62.99 \\
Haiku 4.5 & 0.443 & 3.53 \\
Opus 5 & 6.080 & 48.50 \\
Sonnet 5 & 1.646 & 13.13 \\
DS V3.2 & 0.093 & 0.74 \\
Gem 3.1P & 3.651 & 29.12 \\
Gem FL & 0.595 & 4.75 \\
Gem 3.8F & 0.856 & 6.83 \\
Gemma 31B & 0.055 & 0.44 \\
L4 Mav & 0.076 & 0.61 \\
L4 Scout & 0.048 & 0.39 \\
Mistral M & 0.612 & 4.88 \\
Kimi 2.6 & 0.237 & 1.89 \\
Luna & 0.122 & 0.98 \\
Sol & 1.024 & 8.17 \\
Terra & 1.422 & 11.34 \\
OSS 120B & 0.043 & 0.34 \\
Qwen 235B & 0.050 & 0.40 \\
GLM 5.3 & 0.501 & 4.00 \\
\bottomrule
\end{tabular}

\end{table}

\begin{figure}[tp]
\centering
\includegraphics[width=\columnwidth]{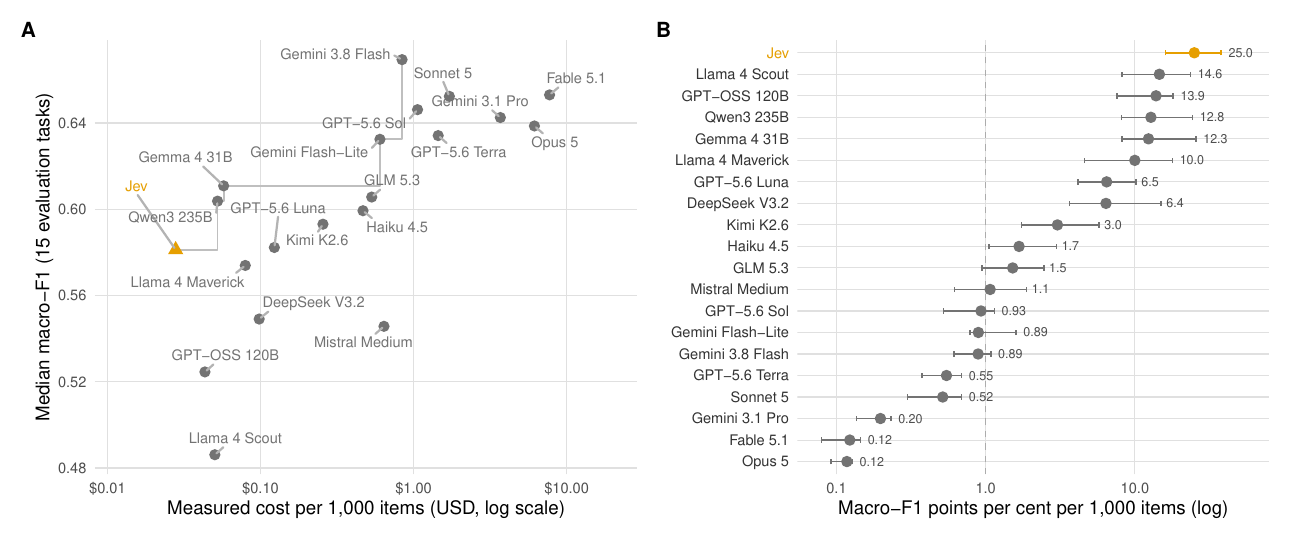}
\caption{Accuracy against measured cost. Panel A places each model by median macro-F1 and median cost per 1{,}000 items over the 15 evaluation tasks (log cost axis; the two local models have no metered cost and appear in Table~\ref{tab:calibration} instead). Panel B shows per-task cost efficiency in macro-F1 points per cent per 1{,}000 items, with dots at the median and whiskers spanning the interquartile range across tasks; the dashed line marks 1 point per cent.}
\label{fig:frontier}
\end{figure}

\begin{figure*}[tp]
\centering
\includegraphics[width=\textwidth]{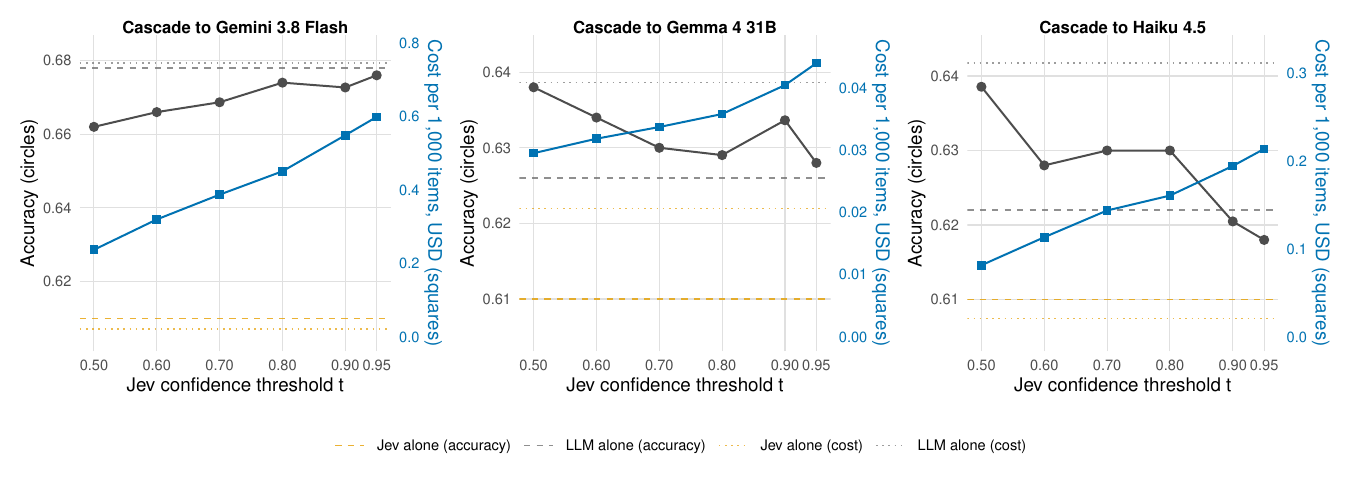}
\caption{Confidence cascades between Jev and three partner LLMs. In each panel, items with Jev confidence at or above the threshold keep Jev's label and the rest are escalated to the partner. Solid lines show median accuracy over the 15 evaluation tasks (left axis, colored by system, dashed reference for the partner alone); dotted lines show median cost (right axis, blue, dotted reference for the partner alone).}
\label{fig:cascade}
\end{figure*}

\subsection{Calibration on dialect text}

Detectors and annotation models routinely perform worst on non-Western and dialectal text, and our dialect-robustness hypothesis expected the same pattern here, namely, that Jev's calibration error on the Indian English dialect task would exceed its median over the other utterance-family evaluation tasks. The concern was that a confidence signal trained on commercial decision workloads would be least trustworthy where its training data are sparsest.

Our data reject the hypothesis in the opposite direction. Jev's calibration error on the dialect task is 0.063, against a median of 0.157 over the other seven utterance-family tasks, and it is the lowest of all 22 models on that task; the next lowest is 0.181. The dialect task is also where Jev's accuracy advantage is largest ($+4.2$ F1 points over the best LLM, the only positive point estimate in the grid) and where its errors are most complementary to LLM errors: among the 266 items, Jev alone is correct on 106 while the best LLM alone is correct on 40. One test on one dialect cannot certify fairness, and we return to this limit in the Discussion, but the specific failure direction we anticipated did not occur.

\subsection{Coverage and accuracy under confidence routing}

We next ask whether the calibrated confidence scores of a decision model could be used to route low-confidence labels to a human or a stronger model. Our routing hypothesis required that, on the median evaluation task, items with Jev confidence at or above 0.9 reach an accuracy of at least 0.85 while covering at least 15 percent of items; the two thresholds set a minimal bar for routing to be worthwhile, requiring confident labels to be accurate enough to keep and frequent enough to reduce annotation cost. Figure~\ref{fig:routing} in the appendix shows the full per-task coverage--accuracy curves.

At the 0.9 threshold, the median task covers 0.376 of items (95 percent CI $[0.202, 0.526]$) at an accuracy of 0.815 (CI $[0.691, 0.897]$). The coverage condition holds while the accuracy condition fails on the point estimate, and the hypothesis is therefore unsupported, though the accuracy interval includes the 0.85 target. Around this median, the task-level spread is wide. Six of the 14 evaluation tasks with a non-empty confident subset clear the 0.85 target (figurative language 0.975, Indian English dialect 0.922, semantic change 0.909), while conversation-level tasks fall well short (toxicity prediction 0.611, Wikipedia corpus power 0.644). The clearest failure is the empathy task, where 78 percent of items receive confidence of at least 0.9 and their accuracy is 0.383 against a full-task base rate of 0.371. Excluding empathy raises the median accuracy at the 0.9 threshold only to 0.824, so the hypothesis fails with or without it. Confident items alone are therefore not accurate enough to keep unconditionally, but the confidence signal retains ranking value on most tasks, and Section~\ref{sec:cascades} below tests whether it is still worth routing on when the escalated items go to an LLM rather than a human.

\subsection{Cost, efficiency, and cascades}
\label{sec:cascades}

The decision-model class separates from every other model in the grid on cost (Figure~\ref{fig:frontier}A). Labeling all 7{,}977 items across the 18 tasks cost \$0.21 with Jev, against \$0.34 to \$62.99 for the 19 LLM baselines; per 1{,}000 items the measured prices are \$0.027 for Jev, \$0.04 to \$0.50 for the hosted open-weight baselines, and \$0.12 to \$7.90 for the closed baselines (Table~\ref{tab:cost}). On the median evaluation task Jev delivers 25.0 macro-F1 points per cent per 1{,}000 items, against 0.59 for the per-task best LLM (Figure~\ref{fig:frontier}B), and the per-task best LLM costs a median 44 times more than Jev. Within Jev's own price neighborhood the comparison narrows: among the six hosted open-weight baselines under \$0.10 per 1{,}000 items, Jev's median macro-F1 of 0.581 exceeds four and trails Gemma 4 31B (0.611) and Qwen3 235B (0.604). On the two tasks where the accuracy confidence interval spans zero under both the fixed and selection-aware comparisons, Jev matches the best LLM at 35 times (ideological books) and 44 times (dialect) lower measured cost; character tropes carries a still larger cost gap of 141 times, but its accuracy parity does not survive the selection-aware comparison (Appendix~\ref{app:exploratory}).

We next ask whether the two model classes are worth combining. Specifically, we route each item by Jev's confidence, keeping Jev's label above the threshold and escalating the remainder to a partner LLM. We compare cascades to two partners across six confidence thresholds: the best frontier LLM, Gemini 3.8 Flash (the highest median macro-F1, 0.669), and the best LLM under \$0.50 per 1{,}000 items, Gemma 4 31B. Figure~\ref{fig:cascade} shows the trade-off. The Gemini cascade holds the partner's accuracy at a fraction of its price: at the 0.8 threshold the median task reaches 0.674 against 0.678 for Gemini alone, at 56 percent of its cost. The Gemma cascade exceeds the partner outright, reaching 0.638 against 0.626 at the 0.5 threshold, because the two systems err on different items. The third panel shows cascades to Haiku 4.5, the best closed model under the same price cap, and shows the same pattern: 0.639 against 0.622 for Haiku alone, at 27 percent of its cost.

\section{Discussion}

Decision models are marketed on three properties: labels at hundredths of a cent, a probability distribution over every label set, and a confidence score for every answer. Computational social science is a natural adopter, because annotation cost has long constrained the field's designs \citep{lazer2009computational} and its published estimates already rest on model-produced labels \citep{gilardi2023chatgpt, tornberg2023chatgpt}. Our evaluation supports the first two properties: labeling the full 7{,}977-item grid cost \$0.21, a median 44 times below the per-task best LLM, and the typed interface eliminated parsing failures, with zero invalid answers across the 23{,}703 decision-model calls of the main evaluation. The third held on some constructs and failed on others, and all three of our hypotheses about the stated uncertainty, that its calibration error would be lower than every baseline's, that high-confidence items would reach 0.85 accuracy at useful coverage, and that calibration would degrade on dialect text, were unsupported.

On accuracy, the decision model is not a substitute for a frontier LLM. It trails the per-task best baseline on 14 of 15 evaluation tasks, by a median of 11.6 macro-F1 points, and none of the five task characteristics we examined (number of answer options, label balance, context length, task family, and difficulty for a fine-tuned baseline) predicted where the gap would close. The comparison to 2023 is instructive in both directions: today's best zero-shot models clear the GPT-4 numbers of \citet{ziems2024llms} on nearly every task, yet fine-tuned RoBERTa-large still wins on half the suite, so the case for zero-shot annotation remains one about convenience and coverage rather than accuracy.

The decision model's native distributions beat the verbalized confidence of 16 of 19 baselines, consistent with a literature that finds verbalized confidence systematically overconfident \citep{xiong2023can, tian2023just}, and three frontier models from a single vendor show a lower median calibration error than the decision model on that same signal. There are two potential explanations, and they are not exclusive. First, the RLCD training distribution is commercial decision workloads, and contested social science constructs are far from it, while frontier post-training appears increasingly to reward accurate self-assessment. Second, verbalized confidence draws on the full capability of a frontier model, so a stronger underlying classifier has less irreducible uncertainty to misstate. An artifact of our elicitation can be ruled out, because every LLM received the same single appended line and the robustness check shows its accuracy effect is negligible. Under either explanation, the 2023-era generalization that language models cannot state trustworthy confidence in words does not hold uniformly on this suite, and verbalized confidence is now a serious competitor to the class's native distributions.

The routing failure is concentrated rather than general. Six of 14 tasks cleared the 0.85 accuracy target in the confident subset, while on empathy classification the model assigned confidence of at least 0.9 to 78 percent of items and was wrong on 62 percent of them. Validating the confidence signal on one task therefore says little about the next, because the tasks where accuracy is lowest are not accompanied by low confidence scores. The dialect test makes the same point in the opposite direction, since we predicted calibration would degrade on Indian English dialect and instead found the model's best-calibrated task. Both of our results support the same practice, a small gold-labeled validation set per construct before any confidence threshold is trusted.

For annotation pipelines, our results support a division of labor rather than a replacement. Used alone, the decision model is a reasonable choice where its accuracy is already competitive, and on those tasks the measured savings are 35 and 44 times (Appendix~\ref{app:exploratory}). Used as the first stage of a confidence cascade, it holds a frontier partner's accuracy at roughly half the cost and exceeds a cheap partner outright, because the two classes err on different items. We note that API prices are intrinsic only for closed models, and for open-weight partners they are hosting prices that a well-utilized local deployment could undercut. For a team with its own hardware, open-weight LLMs label at cost; for one without, hosted open-weight inference already reaches Jev's price neighborhood, and in that bracket the decision model's case rests on its returned distribution and its complementary errors rather than on price alone.

Two properties of decision models nonetheless offer value regardless of where they rank on accuracy. The label space is enforced by construction, so the silent-parse-failure mode of chat-model annotation does not exist; constrained decoding can retrofit much of the same guarantee onto an LLM, though in our structured-output control (Appendix~\ref{app:exploratory}) it moved accuracy by less than a point and still let a handful of invalid answers through where the decisions endpoint allowed none. The returned object is a distribution over the codebook, so calibration is a measurable property of every labeling run rather than a separate elicitation. A model class need not lead on accuracy to be useful in the annotation pipeline if it is inexpensive and explicit about its uncertainty, and our cascade results quantify where and when that combination is useful.
This study is a 2026 snapshot of one commercial decision model, thirteen open-weight counterparts between 0.15B and 9B parameters, and 18 English-language classification tasks, and it does not establish that decision models are calibrated on social science text in general, that the empathy failure is specific to that construct, or that frontier verbalized confidence will remain well calibrated as vendors change post-training recipes. All tasks are public benchmarks that predate the evaluated models, so training-data contamination cannot be excluded, and our results measure performance on established CSS benchmarks rather than generalization to genuinely unseen annotation problems. Three avenues follow directly. First, the headline evaluation covers a single vendor's model at one point in time, and although the open ecosystem in Appendix~\ref{app:exploratory} already spans encoders, decoders, and adapters, all but one of its systems were released within days of Jev; the mirror design is cheap to rerun as the class matures. Second, our dialect-robustness evidence is a single task, and a fuller audit across dialects, languages, and contested constructs is needed before any equity claim. Finally, post-hoc calibration deserves systematic study: in Appendix~\ref{app:exploratory}, a scalar temperature fit on the three pilot tasks moves several inexpensive LLMs ahead of the decision model on evaluation-task calibration, so the class's native-calibration advantage may not survive even minimal tuning of its competitors.

Ziems et al.\ asked whether language models can annotate social science text and answered with a qualified yes. The successor question is no longer the price of the labels, which is settled; it is whether the model's own account of their quality can be trusted. On present evidence it can be on some constructs and cannot be on others, and the two cases cannot be told apart without a per-construct check.

\section{Ethics and Limitations}
All datasets are the public releases of \citet{ziems2024llms} and their upstream sources; no new human data was collected, and all reported quantities are aggregates over public benchmark items. Hate speech and toxicity items are quoted only as required for error analysis. Decision-model confidence scores evaluated here are contestable measurements, and our dialect-robustness cell tests one specific failure direction (non-Western dialect text); it does not certify fairness elsewhere, and the substantive limitations and the avenues they open are discussed at the end of the Discussion. Total inference spend for the full grid was \$244.54, and the complete per-call records and analysis scripts are released for replication at \url{https://github.com/hazemibrahim97/decision-models-css}.

\bibliographystyle{unsrtnat}
\bibliography{refs_expanded}

\appendix

\section{Full results}
\label{app:registered}

\subsection{Complete per-cell results}
Table~\ref{tab:full} reports every cell of the grid: accuracy, macro-F1, expected calibration error, Brier score, and measured cost per 1{,}000 items for all 22 models of the main evaluation on all 18 tasks and for the eleven open decision systems on the tasks they cover, together with the Jev-minus-LLM $\Delta$F1, its paired-bootstrap 95 percent confidence interval, and the BH-FDR-adjusted $q$-value for each of the 342 comparisons (194 significant at $q = 0.05$).

\subsection{Reliability diagrams}
\begin{figure}[!htbp]
\centering
\includegraphics[width=\textwidth]{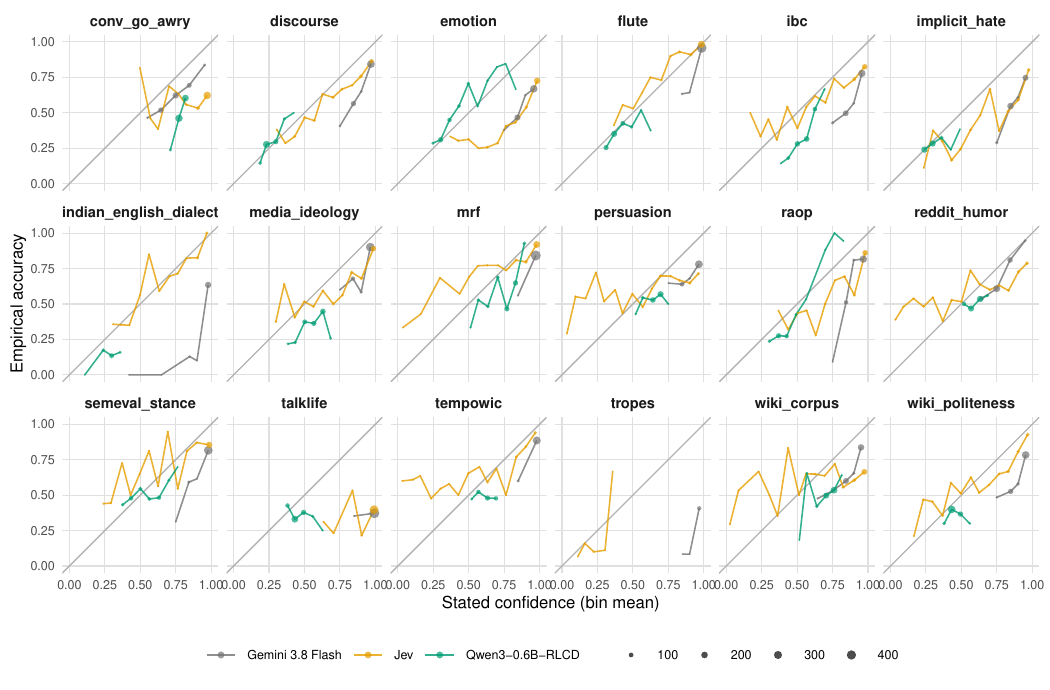}
\caption{Reliability diagrams on all 18 tasks for Jev (orange), Gemini 3.8 Flash (grey, verbalized confidence), and Qwen3-0.6B-RLCD (green). Expected calibration error is computed on all 15 equal-width bins; bins with fewer than 10 items are not displayed, and point size marks the number of items in the bin.}
\label{fig:reliability}
\end{figure}

Figure~\ref{fig:reliability} shows reliability diagrams on all 18 tasks for Jev, the best frontier LLM (Gemini 3.8 Flash, verbalized confidence), and the open-weight Qwen3-0.6B-RLCD decision model. The empathy panel (talklife) shows the failure described in the main text, a flat accuracy profile across the entire confidence range.

\subsection{Routing curves per task}
\begin{figure}[!htbp]
\centering
\includegraphics[width=\textwidth]{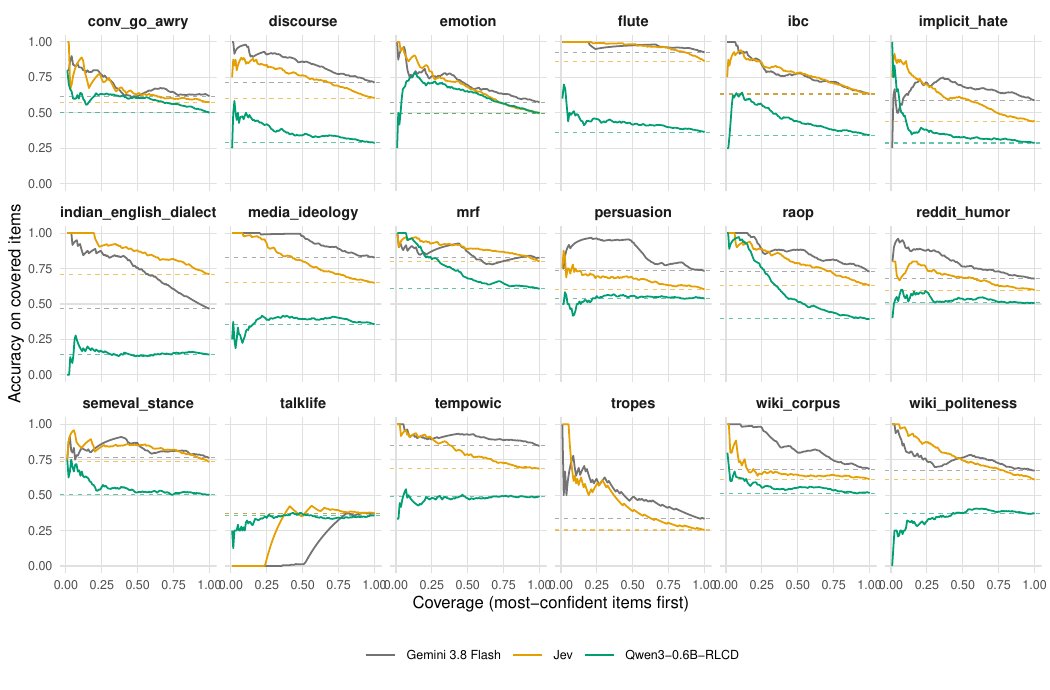}
\caption{Coverage--accuracy routing curves per task for Jev, Gemini 3.8 Flash, and Qwen3-0.6B-RLCD. Each curve traces accuracy within the subset of items at or above a confidence threshold as the threshold rises from 0.5 to 0.95.}
\label{fig:routing}
\end{figure}

\begin{figure}[!htbp]
\centering
\includegraphics[width=\textwidth]{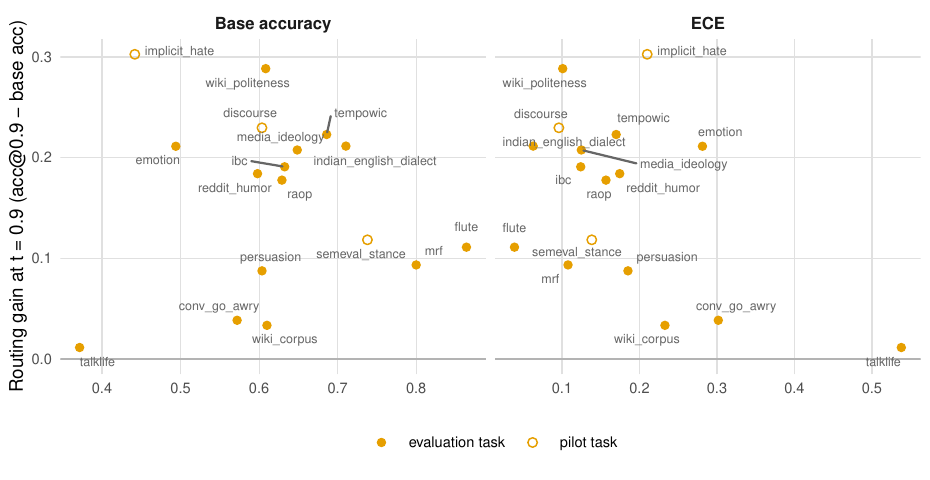}
\caption{Routing gain against task-level calibration error and base accuracy, the descriptive analysis of when confidence routing helps.}
\label{fig:routing_gain}
\end{figure}

Figure~\ref{fig:routing} shows the coverage--accuracy routing curves per task for the same three models, and Figure~\ref{fig:routing_gain} the descriptive analysis relating routing gain to task-level calibration error and base accuracy.

\subsection{Exclusion accounting}
Table~\ref{tab:exclusions} summarizes the exclusion accounting per model, and Table~\ref{tab:exclusions_full} breaks it out by task and model for every cell with a nonzero count. The extraction rules (answer parser and verbalized-confidence pattern) are fixed in code and released with the replication package.

\begin{table}[!htbp]
\centering
\caption{Exclusion accounting per model, summed over tasks: API failures after retries, persistently empty completions, invalid or unparseable answers (excluded from all metrics), and items whose verbalized confidence was absent or negative (kept for accuracy, excluded from calibration and routing). DeepSeek V3.2 emits negative confidence as a refusal-to-estimate convention on 55 percent of items. $^{\dagger}$ marks open decision systems added after the analysis plan was filed (Appendix~\ref{app:exploratory}).}
\label{tab:exclusions}
\begin{tabular}{lrrrrr}
\toprule
Model & API gap & Empty & Invalid & Conf.\ absent & Conf.\ negative \\
\midrule
Jev & 0 & 0 & 0 & 0 & 0 \\
Qwen3-0.6B-RLCD & 0 & 0 & 0 & 0 & 0 \\
Qwen3-0.6B base & 0 & 0 & 0 & 0 & 0 \\
Verdict-0.15B$^{\dagger}$ & 0 & 0 & 0 & 0 & 0 \\
Laya-0.4B$^{\dagger}$ & 0 & 0 & 0 & 0 & 0 \\
NLI-0.4B$^{\dagger}$ & 0 & 0 & 0 & 0 & 0 \\
Von-0.4B$^{\dagger}$ & 0 & 0 & 0 & 0 & 0 \\
decider-0.8b$^{\dagger}$ & 0 & 0 & 0 & 0 & 0 \\
Kev-0.8B$^{\dagger}$ & 0 & 0 & 0 & 0 & 0 \\
decider-2b$^{\dagger}$ & 0 & 0 & 0 & 0 & 0 \\
SemIf-4B$^{\dagger}$ & 0 & 0 & 0 & 0 & 0 \\
Kev-4B$^{\dagger}$ & 0 & 0 & 0 & 0 & 0 \\
Nimble-9B$^{\dagger}$ & 0 & 0 & 0 & 0 & 0 \\
Kev-9B$^{\dagger}$ & 0 & 0 & 0 & 0 & 0 \\
Fable 5.1 & 0 & 14 & 312 & 0 & 0 \\
Haiku 4.5 & 0 & 0 & 375 & 2 & 0 \\
Opus 5 & 0 & 7 & 17 & 1 & 0 \\
Sonnet 5 & 0 & 14 & 137 & 0 & 0 \\
DS V3.2 & 0 & 0 & 267 & 0 & 4389 \\
Gem 3.1P & 1 & 0 & 158 & 0 & 0 \\
Gem FL & 1 & 0 & 80 & 1 & 0 \\
Gem 3.8F & 1 & 0 & 34 & 1 & 0 \\
Gemma 31B & 0 & 0 & 217 & 0 & 0 \\
L4 Mav & 0 & 0 & 132 & 0 & 0 \\
L4 Scout & 0 & 0 & 319 & 0 & 0 \\
Mistral M & 0 & 0 & 392 & 0 & 0 \\
Kimi 2.6 & 0 & 0 & 192 & 1 & 0 \\
Luna & 0 & 0 & 31 & 0 & 0 \\
Sol & 0 & 0 & 9 & 0 & 0 \\
Terra & 0 & 0 & 53 & 7 & 0 \\
OSS 120B & 0 & 0 & 158 & 17 & 0 \\
Qwen 235B & 0 & 0 & 98 & 2 & 0 \\
GLM 5.3 & 0 & 0 & 167 & 5 & 0 \\
\bottomrule
\end{tabular}

\end{table}

\subsection{Reproducibility record}
Table~\ref{tab:repro} reports, for every model, the resolved version returned by the API on each call, the serving providers observed, and the access route. Main-evaluation runs are dated 2026-09-20, with the two 0.6B local models pinned to Hugging Face snapshot hashes; the open decision systems were run 2026-09-20 to 2026-09-23 and are identified by the version string each system's released code reports.

{\footnotesize\begin{longtable}{@{}l >{\raggedright\arraybackslash}p{0.27\textwidth} >{\raggedright\arraybackslash}p{0.34\textwidth} l@{}}
\caption{Reproducibility record, generated from the per-call logs: resolved model version, serving providers observed, and access route. Main-evaluation runs 2026-09-20, with the two 0.6B local models pinned to Hugging Face snapshot hashes. $^{\dagger}$ marks open decision systems added after the analysis plan was filed (Appendix~\ref{app:exploratory}) were run 2026-09-20 to 2026-09-23 and are identified by the version string each system's released code reports.}\label{tab:repro}\\
\toprule
Model & Resolved version(s) & Provider(s) & Route \\
\midrule
\endfirsthead
\toprule
Model & Resolved version(s) & Provider(s) & Route \\
\midrule
\endhead
\bottomrule
\endfoot
Jev & \texttt{jev-1.13-20260917} & -- & decisions API \\
Qwen3-0.6B-RLCD & \texttt{b327ec5efb5f\allowbreak dbf8bfafa3b3\allowbreak 69720ac5f643\allowbreak 4b05} & -- & local (HF) \\
Qwen3-0.6B base & \texttt{Qwen3-0.6B-Base} & -- & local (HF) \\
Verdict-0.15B$^{\dagger}$ & \texttt{heman10x/rlc\allowbreak d-modernbert\allowbreak -151m@v1.4 (\allowbreak artifacts/v2\allowbreak  onnx)} & -- & local (HF) \\
Laya-0.4B$^{\dagger}$ & \texttt{laya-0.3.4/c\allowbreak onvaiinnovat\allowbreak ions/laya\,/\,laya-0.3.4/c\allowbreak onvaiinnovat\allowbreak ions/laya/mu\allowbreak ltilingual} & -- & local (HF) \\
NLI-0.4B$^{\dagger}$ & \texttt{opendecision\allowbreak -0.1.1/Morit\allowbreak zLaurer/Mode\allowbreak rnBERT-large\allowbreak -zeroshot-v2\allowbreak .0} & -- & local (HF) \\
Von-0.4B$^{\dagger}$ & \texttt{wfzyx/von we\allowbreak ights-1.1.0 \allowbreak snapshot-d8b\allowbreak b5e0 sdk-1.0\allowbreak .1} & -- & local (HF) \\
decider-0.8b$^{\dagger}$ & \texttt{Mapika/decider-0.8b} & -- & local (HF) \\
Kev-0.8B$^{\dagger}$ & \texttt{jaredpalmer/kev-0.8b} & -- & local (HF) \\
decider-2b$^{\dagger}$ & \texttt{Mapika/decider-2b} & -- & local (HF) \\
SemIf-4B$^{\dagger}$ & \texttt{semif-direct\allowbreak -mlx/Qwen/Qw\allowbreak en3.5-4B@851\allowbreak bf6e8} & -- & local (HF) \\
Kev-4B$^{\dagger}$ & \texttt{jaredpalmer/kev-4b} & -- & local (HF) \\
Nimble-9B$^{\dagger}$ & \texttt{bespokelabs/\allowbreak Bespoke-Nimb\allowbreak le-9B} & -- & local (HF) \\
Kev-9B$^{\dagger}$ & \texttt{jaredpalmer/kev-9b} & -- & local (HF) \\
Fable 5.1 & \texttt{claude-fable-5.1} & Anthropic, Google & chat completions \\
Haiku 4.5 & \texttt{claude-haiku-4.5} & Amazon Bedrock, Anthropic & chat completions \\
Opus 5 & \texttt{claude-opus-5} & Claude Platform on AWS & chat completions \\
Sonnet 5 & \texttt{claude-sonnet-5} & Claude Platform on AWS & chat completions \\
DS V3.2 & \texttt{deepseek-v3.2} & Alibaba, AtlasCloud, Baidu, DeepInfra, DigitalOcean, Friendli, GMICloud, Google, Novita, Phala, SiliconFlow, StreamLake, Venice & chat completions \\
Gem 3.1P & \texttt{gemini-3.1-pro-preview} & Google, Google AI Studio & chat completions \\
Gem FL & \texttt{gemini-3.5-flash-lite} & Google, Google AI Studio & chat completions \\
Gem 3.8F & \texttt{gemini-3.8-flash} & Google, Google AI Studio & chat completions \\
Gemma 31B & \texttt{gemma-4-31b-it} & Chutes, CoreWeave, Crusoe, DeepInfra, Friendli, ModelRun, Novita, Parasail, SambaNova, Venice & chat completions \\
L4 Mav & \texttt{llama-4-maverick} & DeepInfra, DigitalOcean, Novita, Parasail & chat completions \\
L4 Scout & \texttt{llama-4-scout} & DeepInfra, Novita & chat completions \\
Mistral M & \texttt{mistral-medium-3-5} & Mistral & chat completions \\
Kimi 2.6 & \texttt{kimi-k2.6} & AtlasCloud, Baidu, Chutes, Cloudflare, CoreWeave, Crusoe, Decart, DeepInfra, DigitalOcean, GMICloud, Inceptron, Moonshot AI, Novita, Parasail, Phala, Sail Research, SiliconFlow, StreamLake, Venice & chat completions \\
Luna & \texttt{gpt-5.6-luna} & Azure, OpenAI & chat completions \\
Sol & \texttt{gpt-5.6-sol} & Azure, OpenAI & chat completions \\
Terra & \texttt{gpt-5.6-terra} & Azure, OpenAI & chat completions \\
OSS 120B & \texttt{gpt-oss-120b} & AkashML, Amazon Bedrock, BaseTen, Cerebras, CoreWeave, Crusoe, DeepInfra, DekaLLM, DigitalOcean, Google, Groq, Mancer 2, Mara, Nebius, Novita, Parasail, Phala, SambaNova, SiliconFlow, Together & chat completions \\
Qwen 235B & \texttt{qwen3-235b-a22b-2507} & Alibaba, DeepInfra, GMICloud, Google, Nebius, Novita, Parasail, StreamLake, Venice & chat completions \\
GLM 5.3 & \texttt{glm-5.3} & AkashML, Alibaba, AtlasCloud, Baidu, BaseTen, Cloudflare, Crusoe, Decart, DeepInfra, DigitalOcean, Fireworks, Friendli, GMICloud, Inceptron, InferenceNet, Io Net, Makora, Mistral, Modal, Morph, Novita, Parasail, Phala, Reka, Sail Research, SiliconFlow, Together, Venice, Wafer, Z.AI & chat completions \\
\end{longtable}
}

\subsection{Confidence-line robustness check}
The robustness check compares each LLM's accuracy on the three pilot tasks with and without the appended confidence-elicitation line. Table~\ref{tab:confline} reports all 48 cells: the median change is $+0.2$ accuracy points, the median absolute change 1.7 points, and the largest 11.4 points (Haiku 4.5 on implicit hate), so the elicitation line does not systematically help or hurt the accuracy comparisons.

\begin{table}[!htbp]
\centering
\caption{Robustness check: accuracy change (points, with-confidence minus without) per LLM on the three pilot tasks when the confidence-elicitation line is appended to the prompt.}
\label{tab:confline}
\small
\begin{tabular}{lrrr}
\toprule
Model & Discourse & Impl. Hate & Stance \\
\midrule
DS V3.2 & -1.0 & -6.2 & -6.7 \\
Fable 5.1 & +0.4 & +3.6 & -2.1 \\
GLM 5.3 & +0.2 & +11.4 & -2.5 \\
Gemma 31B & +0.2 & -0.2 & +0.5 \\
Haiku 4.5 & -0.4 & -2.6 & +0.2 \\
Kimi 2.6 & +6.6 & +9.2 & -1.1 \\
L4 Mav & +2.8 & +5.8 & +1.6 \\
L4 Scout & -0.2 & +7.8 & -5.5 \\
Luna & +0.0 & +2.2 & +1.4 \\
Mistral M & +0.6 & +3.8 & -4.4 \\
OSS 120B & -2.6 & +4.6 & +0.9 \\
Opus 5 & +0.0 & +1.2 & +1.8 \\
Qwen 235B & +0.8 & +4.6 & -1.1 \\
Sol & +0.2 & -0.4 & +0.2 \\
Sonnet 5 & +1.0 & -4.0 & -2.3 \\
Terra & +1.8 & +0.2 & -2.5 \\
\bottomrule
\end{tabular}

\end{table}

\FloatBarrier
{\footnotesize\setlength{\tabcolsep}{3.5pt}\begin{longtable}{llrrrrrlr}
\caption{Complete per-cell results: accuracy, macro-F1, ECE (15 bins), Brier, measured \$/1{,}000 items, and, for each LLM, the Jev-minus-LLM $\Delta$F1 (points) with paired-bootstrap 95\% CI and BH-FDR $q$ ($^{*}$: significant at $q=0.05$). $^{d}$ marks pilot tasks; -- marks the tropes cells unrunnable for the 0.6B models and cells without a Jev comparison; $^{\dagger}$ marks open decision systems added after the analysis plan was filed (Appendix~\ref{app:exploratory}), which run locally at no API cost and cover 17 tasks (SemIf 16).}\label{tab:full}\\
\toprule
Task & Model & Acc & F1 & ECE & Brier & \$/1k & $\Delta$F1 [95\% CI] & $q$ \\
\midrule
\endfirsthead
\toprule
Task & Model & Acc & F1 & ECE & Brier & \$/1k & $\Delta$F1 [95\% CI] & $q$ \\
\midrule
\endhead
\bottomrule
\endfoot
Dialect & Jev & 0.711 & 0.640 & 0.063 & 0.425 & 0.048 & -- & -- \\
 & Qwen3-0.6B-RLCD & 0.143 & 0.057 & 0.181 & 0.936 & 0.000 & -- & -- \\
 & Qwen3-0.6B base & 0.440 & 0.027 & 0.255 & 0.863 & 0.000 & -- & -- \\
 & Verdict-0.15B$^{\dagger}$ & 0.019 & 0.002 & 0.086 & 0.977 & 0.000 & -- & -- \\
 & Laya-0.4B$^{\dagger}$ & 0.425 & 0.046 & 0.562 & 1.132 & 0.000 & -- & -- \\
 & NLI-0.4B$^{\dagger}$ & 0.244 & 0.234 & 0.140 & 0.908 & 0.000 & -- & -- \\
 & Von-0.4B$^{\dagger}$ & 0.256 & 0.086 & 0.351 & 1.173 & 0.000 & -- & -- \\
 & decider-0.8b$^{\dagger}$ & 0.188 & 0.265 & 0.113 & 0.952 & 0.000 & -- & -- \\
\end{longtable}
}

{\small\begin{longtable}{llrrrrrr}
\caption{Per task and model exclusion counts, all cells with a nonzero count (columns as in Table~\ref{tab:exclusions}; C.\ = confidence).}\label{tab:exclusions_full}\\
\toprule
Task & Model & API & Empty & Invalid & C.\ abs & C.\ neg & C.\ oob \\
\midrule
\endfirsthead
\toprule
Task & Model & API & Empty & Invalid & C.\ abs & C.\ neg & C.\ oob \\
\midrule
\endhead
\bottomrule
\endfoot
Dialect & DS V3.2 & 0 & 0 & 0 & 0 & 162 & 0 \\
Dialect & Fable 5.1 & 0 & 0 & 45 & 0 & 0 & 0 \\
Dialect & GLM 5.3 & 0 & 0 & 17 & 0 & 0 & 0 \\
Dialect & Gem 3.1P & 0 & 0 & 2 & 0 & 0 & 0 \\
Dialect & Gem 3.8F & 0 & 0 & 2 & 0 & 0 & 0 \\
Dialect & Gem FL & 0 & 0 & 10 & 0 & 0 & 0 \\
Dialect & Gemma 31B & 0 & 0 & 7 & 0 & 0 & 0 \\
Dialect & Haiku 4.5 & 0 & 0 & 9 & 0 & 0 & 0 \\
\end{longtable}
}

\clearpage
\section{Exploratory analyses}
\label{app:exploratory}

\paragraph{ECE binning sensitivity.}
Table~\ref{tab:b1} recomputes each model's median evaluation-task ECE under 10, 15, and 20 equal-width bins and under 15 equal-mass bins. The model ordering is stable across binning choices. Equal-mass binning raises the ECE of the LLMs with strongly discretized verbalized confidence (Fable 5.1 moves from 0.073 to 0.123, Sonnet 5 from 0.075 to 0.156) while leaving Jev's quasi-continuous distribution nearly unchanged (0.157 to 0.154), so part of the frontier verbalized-confidence advantage in the primary 15-bin metric reflects confidence values concentrated on a few round numbers; Opus 5 remains below Jev under every binning (0.089 equal-mass).

\begin{table}[!htbp]
\centering
\caption{Median evaluation-task ECE under alternative binning schemes; 15 equal-width bins is the primary metric.}
\label{tab:b1}
\small
\begin{tabular}{lrrrr}
\toprule
Model & 10 bins & 15 bins (primary) & 20 bins & 15 equal-mass \\
\midrule
Opus 5 & 0.0549 & 0.0664 & 0.0692 & 0.0889 \\
Fable 5.1 & 0.0698 & 0.0733 & 0.0746 & 0.1230 \\
Sonnet 5 & 0.0629 & 0.0752 & 0.0752 & 0.1558 \\
Qwen3-0.6B-RLCD & 0.1278 & 0.1283 & 0.1283 & 0.1316 \\
Qwen3-0.6B base & 0.1414 & 0.1443 & 0.1438 & 0.1509 \\
Jev & 0.1567 & 0.1567 & 0.1600 & 0.1539 \\
Haiku 4.5 & 0.1648 & 0.1668 & 0.1668 & 0.2087 \\
GLM 5.3 & 0.1668 & 0.1678 & 0.1678 & 0.1955 \\
Gem 3.8F & 0.1889 & 0.1889 & 0.1907 & 0.2018 \\
Kimi 2.6 & 0.1859 & 0.1921 & 0.1937 & 0.2196 \\
DS V3.2 & 0.1937 & 0.1937 & 0.1983 & 0.2690 \\
Gem FL & 0.2145 & 0.2150 & 0.2150 & 0.2391 \\
Terra & 0.2158 & 0.2158 & 0.2158 & 0.2158 \\
Gem 3.1P & 0.2326 & 0.2326 & 0.2326 & 0.2362 \\
Sol & 0.2334 & 0.2334 & 0.2350 & 0.2334 \\
Mistral M & 0.2608 & 0.2608 & 0.2608 & 0.2821 \\
L4 Mav & 0.2685 & 0.2713 & 0.2713 & 0.2887 \\
Gemma 31B & 0.2854 & 0.2854 & 0.2854 & 0.2933 \\
OSS 120B & 0.2867 & 0.2869 & 0.2887 & 0.2873 \\
Qwen 235B & 0.3074 & 0.3074 & 0.3074 & 0.3246 \\
L4 Scout & 0.3122 & 0.3122 & 0.3122 & 0.3337 \\
Luna & 0.3346 & 0.3335 & 0.3359 & 0.3310 \\
\bottomrule
\end{tabular}

\end{table}

\paragraph{Post-hoc temperature scaling.}
For each model we fit a scalar temperature on the logit of stated confidence by maximum likelihood on the pooled pilot-task items, then recompute evaluation-task ECE with the scaled confidence (Table~\ref{tab:b2}). Scaling helps most where miscalibration is largest: Llama 4 Scout improves from 0.312 to 0.060 and Gemini 3.1 Pro from 0.233 to 0.076, both moving ahead of Jev, whose own scaled ECE is 0.121 (from 0.157, $T = 2.66$). Already-calibrated models gain nothing (Fable 5.1 unchanged) or lose slightly in the transfer across task sets (Opus 5, 0.066 to 0.086). A small labeled calibration set therefore yields frontier-level calibration for several inexpensive LLMs, which weakens the case that native calibration is the class's advantage.

\begin{table}[!htbp]
\centering
\caption{Scalar temperature $T$ fit on the three pilot tasks and median evaluation-task ECE before and after scaling.}
\label{tab:b2}
\small
\begin{tabular}{lrrr}
\toprule
Model & $T$ & ECE before & ECE after \\
\midrule
L4 Scout & 8.414 & 0.3122 & 0.0602 \\
Fable 5.1 & 1.413 & 0.0733 & 0.0732 \\
L4 Mav & 6.683 & 0.2713 & 0.0750 \\
Gem 3.1P & 3.162 & 0.2326 & 0.0762 \\
Gem 3.8F & 3.350 & 0.1889 & 0.0834 \\
Qwen 235B & 6.310 & 0.3074 & 0.0846 \\
Opus 5 & 1.585 & 0.0664 & 0.0862 \\
Gem FL & 5.957 & 0.2150 & 0.0888 \\
Sonnet 5 & 1.413 & 0.0752 & 0.0895 \\
OSS 120B & 4.467 & 0.2869 & 0.0956 \\
Terra & 3.758 & 0.2158 & 0.0964 \\
Gemma 31B & 6.310 & 0.2854 & 0.1064 \\
Kimi 2.6 & 2.985 & 0.1921 & 0.1074 \\
DS V3.2 & 5.957 & 0.1937 & 0.1137 \\
Mistral M & 5.623 & 0.2608 & 0.1176 \\
Jev & 2.661 & 0.1567 & 0.1208 \\
Qwen3-0.6B-RLCD & 1.122 & 0.1283 & 0.1234 \\
Sol & 3.758 & 0.2334 & 0.1269 \\
Haiku 4.5 & 2.113 & 0.1668 & 0.1323 \\
Luna & 5.957 & 0.3335 & 0.1330 \\
GLM 5.3 & 2.661 & 0.1678 & 0.1372 \\
Qwen3-0.6B base & 0.750 & 0.1443 & 0.1774 \\
\bottomrule
\end{tabular}

\end{table}

\paragraph{Option-position bias.}
Table~\ref{tab:b3} compares each model's distribution over option positions with the gold distribution on the multi-choice evaluation tasks (first-option share and total variation distance). The two 0.6B models show the largest skews (the untrained base picks the first option on 49 percent of items against a 30 percent gold share; the RLCD model under-picks it at 17 percent); the API models sit in a 0.08--0.17 total-variation band, and Jev shows no distinctive first-option bias.

\begin{table}[!htbp]
\centering
\caption{Option-position bias on the multi-choice evaluation tasks; TV dist.\ is the total variation distance between the predicted and gold position distributions.}
\label{tab:b3}
\small
\begin{tabular}{lrrrr}
\toprule
Model & $n$ & First opt.\ (pred) & First opt.\ (gold) & TV dist. \\
\midrule
Qwen3-0.6B base & 4001 & 0.4879 & 0.2999 & 0.2559 \\
Qwen3-0.6B-RLCD & 4001 & 0.1722 & 0.2999 & 0.2214 \\
Gem FL & 4088 & 0.2233 & 0.2931 & 0.1659 \\
Gemma 31B & 4058 & 0.2070 & 0.2960 & 0.1619 \\
Qwen 235B & 4103 & 0.2391 & 0.2927 & 0.1470 \\
L4 Mav & 4110 & 0.2538 & 0.2922 & 0.1389 \\
Gem 3.1P & 4077 & 0.2394 & 0.2938 & 0.1359 \\
Gem 3.8F & 4104 & 0.2354 & 0.2924 & 0.1318 \\
Kimi 2.6 & 4076 & 0.2230 & 0.2947 & 0.1303 \\
DS V3.2 & 4081 & 0.2460 & 0.2940 & 0.1296 \\
Sol & 4115 & 0.2277 & 0.2919 & 0.1295 \\
Fable 5.1 & 3928 & 0.2360 & 0.2925 & 0.1288 \\
Mistral M & 4091 & 0.2410 & 0.2936 & 0.1183 \\
Terra & 4115 & 0.2292 & 0.2919 & 0.1159 \\
Haiku 4.5 & 4011 & 0.2359 & 0.2992 & 0.1157 \\
Luna & 4114 & 0.2304 & 0.2919 & 0.1108 \\
Jev & 4115 & 0.2433 & 0.2919 & 0.1094 \\
Opus 5 & 4102 & 0.2604 & 0.2925 & 0.1085 \\
Sonnet 5 & 4082 & 0.2482 & 0.2937 & 0.1075 \\
GLM 5.3 & 4054 & 0.2679 & 0.2938 & 0.1053 \\
OSS 120B & 4085 & 0.2874 & 0.2928 & 0.0960 \\
L4 Scout & 4082 & 0.2749 & 0.2940 & 0.0804 \\
\bottomrule
\end{tabular}

\end{table}

\paragraph{Jev confidence field against top-option probability.}
Jev returns both a scalar confidence and a full distribution. Table~\ref{tab:b4} compares per-task ECE computed on each: the two signals are close everywhere (medians 0.157 against 0.168, confidence slightly better), so no headline result depends on which is used.

\begin{table}[!htbp]
\centering
\caption{Per-task ECE of Jev's scalar confidence field against the probability of its chosen option.}
\label{tab:b4}
\small
\begin{tabular}{lrr}
\toprule
Task & ECE (confidence) & ECE ($p_{\mathrm{top}}$) \\
\midrule
Toxicity & 0.3017 & 0.3424 \\
Emotion & 0.2812 & 0.3211 \\
Figurative & 0.0387 & 0.0323 \\
Ideology (utt.) & 0.1242 & 0.1680 \\
Dialect & 0.0628 & 0.0630 \\
Ideology (doc.) & 0.1250 & 0.1637 \\
Misinfo & 0.1077 & 0.0898 \\
Persuasion (conv.) & 0.1851 & 0.1920 \\
Persuasion (utt.) & 0.1567 & 0.1882 \\
Humor & 0.1745 & 0.1912 \\
Empathy & 0.5379 & 0.5653 \\
Sem. Change & 0.1699 & 0.1261 \\
Tropes & 0.1265 & 0.1246 \\
Power & 0.2328 & 0.2484 \\
Politeness & 0.1009 & 0.1388 \\
median & 0.1567 & 0.1680 \\
\bottomrule
\end{tabular}

\end{table}

\paragraph{Confusion structure on Jev's weak tasks.}
Tables~\ref{tab:b5a}--\ref{tab:b5d} show where Jev's errors concentrate on its four weakest tasks. On empathy (talklife) it collapses onto the no-exploration class regardless of the gold level (recall 0.92 on class C, 0.02 on the strong-exploration class A; Table~\ref{tab:b5a}); toxicity errors are near-symmetric (Table~\ref{tab:b5b}); implicit-hate errors concentrate in the stereotypical category, with threatening items most often misread as incitement (Table~\ref{tab:b5c}); and the tropes errors scatter without a dominant confusion pair (Table~\ref{tab:b5d}).

\begin{table}[!htbp]
\centering
\caption{Jev confusion matrix on the empathy task (A = strong exploration, B = weak, C = none).}
\label{tab:b5a}
\small
\begin{tabular}{lrrr}
\toprule
gold $\to$  & A & B & C \\
\midrule
A & 4 & 26 & 136 \\
B & 5 & 28 & 133 \\
C & 0 & 13 & 153 \\
\bottomrule
\end{tabular}

\end{table}

\begin{table}[!htbp]
\centering
\caption{Jev confusion matrix on toxicity prediction.}
\label{tab:b5b}
\small
\begin{tabular}{lrr}
\toprule
gold $\to$  & False & True \\
\midrule
False & 237 & 13 \\
True & 201 & 49 \\
\bottomrule
\end{tabular}

\end{table}

\begin{table}[!htbp]
\centering
\caption{Jev confusion matrix on implicit hate (pilot task).}
\label{tab:b5c}
\small
\resizebox{\columnwidth}{!}{\begin{tabular}{lrrrrrr}
\toprule
gold $\to$  & incitement & inferiority & irony & stereotypical & threatening & white\_grievance \\
\midrule
incitement & 23 & 11 & 10 & 22 & 2 & 15 \\
inferiority & 5 & 29 & 9 & 36 & 1 & 3 \\
irony & 1 & 10 & 39 & 28 & 2 & 3 \\
stereotypical & 6 & 2 & 4 & 58 & 0 & 13 \\
threatening & 35 & 4 & 1 & 18 & 23 & 2 \\
white\_grievance & 5 & 2 & 5 & 20 & 3 & 48 \\
\bottomrule
\end{tabular}
}
\end{table}

\begin{table}[!htbp]
\centering
\caption{The ten most frequent Jev confusions on character tropes.}
\label{tab:b5d}
\small
\begin{tabular}{lrr}
\toprule
Gold trope & Predicted trope & Count \\
\midrule
AA, BC & AA & 1 \\
AE & J & 1 \\
AG, AR & BE & 1 \\
AG & AV, N, BQ & 1 \\
AB & J & 1 \\
AH & BF & 1 \\
AI & BQ & 1 \\
AH, AY & Q & 1 \\
AJ, O, C, K & BD & 1 \\
AJ & V, N, K, AO, I & 1 \\
\bottomrule
\end{tabular}

\end{table}

\paragraph{Item-level comparison with the 2023 evaluation.}
Ziems et al.\ released ChatGPT's per-item answers, which we align to our items by their released row order and score with the same gold labels; alignment verifies exactly on 14 of the 18 tasks (the released humor and tropes files do not match the prompt order and are excluded, and two tasks have no released file). Table~\ref{tab:b6} cross-tabulates per-item correctness. Aggregate accuracy rose from 0.49 (ChatGPT 2023) to 0.59 for Jev and 0.66 for Gemini 3.8 Flash on the same items, and the error sets moved rather than nested: 811 items that ChatGPT solved in 2023 are missed by Jev, against 1{,}434 gained.

\begin{table}[!htbp]
\centering
\caption{Per-item correctness against the released 2023 ChatGPT answers on the 14 alignment-verified tasks.}
\label{tab:b6}
\small
\begin{tabular}{lrrrrrr}
\toprule
Task & Model & $n$ & Both & Only 2026 & Only 2023 & Neither \\
\midrule
Dialect & Gem 3.8F & 264 & 38 & 85 & 4 & 137 \\
Dialect & Jev & 266 & 38 & 151 & 4 & 73 \\
Discourse & Gem 3.8F & 497 & 182 & 172 & 39 & 104 \\
Discourse & Jev & 497 & 173 & 127 & 48 & 149 \\
Emotion & Gem 3.8F & 498 & 196 & 89 & 34 & 179 \\
Emotion & Jev & 498 & 192 & 54 & 38 & 214 \\
Empathy & Gem 3.8F & 498 & 115 & 68 & 72 & 243 \\
Empathy & Jev & 498 & 118 & 67 & 69 & 244 \\
Ideology (doc.) & Gem 3.8F & 498 & 280 & 131 & 13 & 74 \\
Ideology (doc.) & Jev & 498 & 261 & 62 & 32 & 143 \\
Ideology (utt.) & Gem 3.8F & 498 & 194 & 121 & 79 & 104 \\
Ideology (utt.) & Jev & 498 & 205 & 110 & 68 & 115 \\
Impl. Hate & Gem 3.8F & 474 & 82 & 197 & 60 & 135 \\
Impl. Hate & Jev & 498 & 75 & 145 & 73 & 205 \\
Persuasion (conv.) & Gem 3.8F & 434 & 166 & 153 & 64 & 51 \\
Persuasion (conv.) & Jev & 434 & 131 & 131 & 99 & 73 \\
Persuasion (utt.) & Gem 3.8F & 399 & 146 & 145 & 17 & 91 \\
Persuasion (utt.) & Jev & 399 & 155 & 96 & 8 & 140 \\
Politeness & Gem 3.8F & 498 & 168 & 166 & 85 & 79 \\
Politeness & Jev & 498 & 140 & 163 & 113 & 82 \\
Power & Gem 3.8F & 500 & 212 & 130 & 96 & 62 \\
Power & Jev & 500 & 197 & 108 & 111 & 84 \\
Sem. Change & Gem 3.8F & 344 & 172 & 120 & 21 & 31 \\
Sem. Change & Jev & 344 & 136 & 100 & 57 & 51 \\
Stance & Gem 3.8F & 435 & 274 & 58 & 39 & 64 \\
Stance & Jev & 435 & 273 & 48 & 40 & 74 \\
Toxicity & Gem 3.8F & 500 & 169 & 140 & 96 & 95 \\
Toxicity & Jev & 500 & 214 & 72 & 51 & 163 \\
\bottomrule
\end{tabular}

\end{table}

\begin{figure}[!htbp]
\centering
\includegraphics[width=\textwidth]{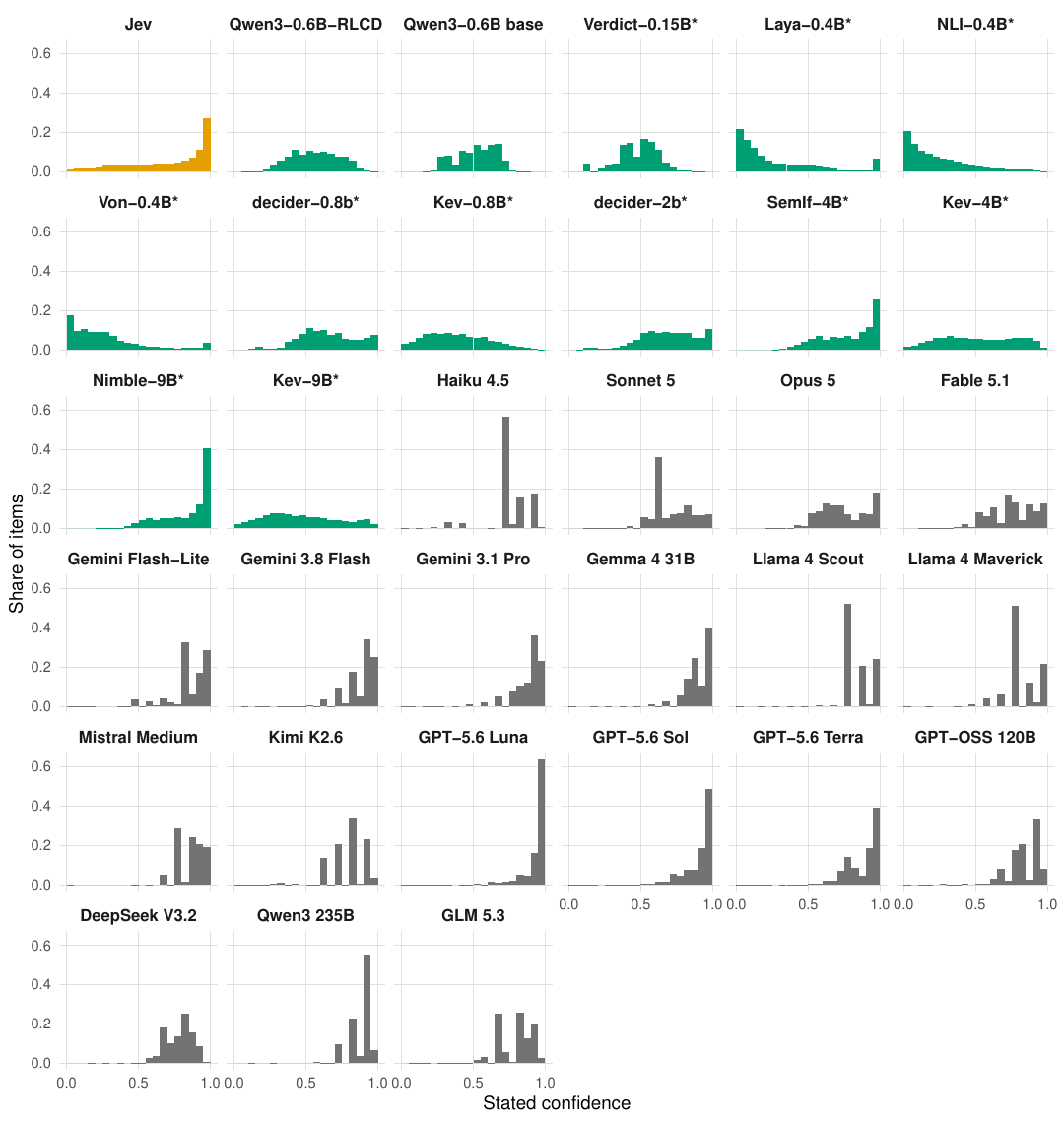}
\caption{Distribution of stated confidence per model, pooled over the evaluation tasks (decision models: returned confidence; LLMs: verbalized confidence). Orange marks Jev, green the other decision models, grey the LLMs; panels marked * are open decision systems added after the analysis plan was filed (Appendix~\ref{app:exploratory}), scored on 14 evaluation tasks (13 for SemIf).}
\label{fig:conf_hist}
\end{figure}

\paragraph{Confidence distributions.}
Figure~\ref{fig:conf_hist} shows each model's stated-confidence distribution pooled over the evaluation tasks. The LLM baselines concentrate on a few round values in the 0.7--0.95 range, Jev spreads mass across the full range with a mode near 1, and the two 0.6B models center near 0.5. Among the open systems, Laya, NLI-0.4B, and Von place about half their items below 0.2 (47 to 58 percent), while SemIf and Nimble place most above 0.8 (52 and 66 percent). These systems define the confidence field differently (Von's is the gap between its two highest probabilities), so the spread reflects the field's definition as much as the model's certainty.

\paragraph{Cascade sensitivity.}
Table~\ref{tab:b8a} recomputes the cascade cost fractions of Figure~\ref{fig:cascade} with the Jev or partner price doubled or halved: the frontier-partner conclusion is insensitive (cost fraction 0.54--0.61 at the 0.8 threshold under every scenario), while the cheap-partner cascade's cost advantage disappears when Jev's price doubles, as expected when the two systems' prices are within a factor of three. Table~\ref{tab:b8b} prices the alternative architecture in which low-confidence items go to a human reviewer instead of an LLM: at the 0.9 threshold the median task reaches 0.944 accuracy under the assumption of perfect human labels, at \$31 to \$625 per 1{,}000 items depending on the assumed human cost of \$0.05 to \$1 per item, against \$0.027 fully automated.

\begin{table}[!htbp]
\centering
\caption{Median cascade cost as a fraction of the partner-alone cost under $\pm 2\times$ price assumptions.}
\label{tab:b8a}
\small
\begin{tabular}{lrrr}
\toprule
Partner & $t$ & Scenario & Median cost fraction \\
\midrule
Haiku 4.5 & 0.5 & baseline & 0.2658 \\
Haiku 4.5 & 0.5 & jev\_x2 & 0.3447 \\
Haiku 4.5 & 0.5 & jev\_x0.5 & 0.2432 \\
Haiku 4.5 & 0.5 & llm\_x2 & 0.2432 \\
Haiku 4.5 & 0.5 & llm\_x0.5 & 0.3447 \\
Haiku 4.5 & 0.8 & baseline & 0.5283 \\
Haiku 4.5 & 0.8 & jev\_x2 & 0.6136 \\
Haiku 4.5 & 0.8 & jev\_x0.5 & 0.4969 \\
Haiku 4.5 & 0.8 & llm\_x2 & 0.4969 \\
Haiku 4.5 & 0.8 & llm\_x0.5 & 0.6136 \\
Haiku 4.5 & 0.95 & baseline & 0.7549 \\
Haiku 4.5 & 0.95 & jev\_x2 & 0.8160 \\
Haiku 4.5 & 0.95 & jev\_x0.5 & 0.7244 \\
Haiku 4.5 & 0.95 & llm\_x2 & 0.7244 \\
Haiku 4.5 & 0.95 & llm\_x0.5 & 0.8160 \\
Gem 3.8F & 0.5 & baseline & 0.2717 \\
Gem 3.8F & 0.5 & jev\_x2 & 0.3017 \\
Gem 3.8F & 0.5 & jev\_x0.5 & 0.2567 \\
Gem 3.8F & 0.5 & llm\_x2 & 0.2567 \\
Gem 3.8F & 0.5 & llm\_x0.5 & 0.3017 \\
Gem 3.8F & 0.8 & baseline & 0.5610 \\
Gem 3.8F & 0.8 & jev\_x2 & 0.6095 \\
Gem 3.8F & 0.8 & jev\_x0.5 & 0.5367 \\
Gem 3.8F & 0.8 & llm\_x2 & 0.5367 \\
Gem 3.8F & 0.8 & llm\_x0.5 & 0.6095 \\
Gem 3.8F & 0.95 & baseline & 0.8057 \\
Gem 3.8F & 0.95 & jev\_x2 & 0.8244 \\
Gem 3.8F & 0.95 & jev\_x0.5 & 0.7963 \\
Gem 3.8F & 0.95 & llm\_x2 & 0.7963 \\
Gem 3.8F & 0.95 & llm\_x0.5 & 0.8244 \\
Gemma 31B & 0.5 & baseline & 0.8443 \\
Gemma 31B & 0.5 & jev\_x2 & 1.4701 \\
Gemma 31B & 0.5 & jev\_x0.5 & 0.5265 \\
Gemma 31B & 0.5 & llm\_x2 & 0.5265 \\
Gemma 31B & 0.5 & llm\_x0.5 & 1.4701 \\
Gemma 31B & 0.8 & baseline & 1.1299 \\
Gemma 31B & 0.8 & jev\_x2 & 1.6934 \\
Gemma 31B & 0.8 & jev\_x0.5 & 0.7939 \\
Gemma 31B & 0.8 & llm\_x2 & 0.7939 \\
Gemma 31B & 0.8 & llm\_x0.5 & 1.6934 \\
Gemma 31B & 0.95 & baseline & 1.3612 \\
Gemma 31B & 0.95 & jev\_x2 & 1.9105 \\
Gemma 31B & 0.95 & jev\_x0.5 & 0.9982 \\
Gemma 31B & 0.95 & llm\_x2 & 0.9982 \\
Gemma 31B & 0.95 & llm\_x0.5 & 1.9105 \\
\bottomrule
\end{tabular}

\end{table}

\begin{table}[!htbp]
\centering
\caption{Jev-plus-human-review pricing; items below the confidence threshold are routed to a human assumed correct, at the stated cost per item.}
\label{tab:b8b}
\small
\begin{tabular}{lrrr}
\toprule
$t$ & \$/item human & Median accuracy & Median \$/1k \\
\midrule
0.9 & 0.05 & 0.9440 & 31.24 \\
0.9 & 0.25 & 0.9440 & 156.14 \\
0.9 & 1.0 & 0.9440 & 624.51 \\
0.95 & 0.05 & 0.9719 & 36.81 \\
0.95 & 0.25 & 0.9719 & 183.80 \\
0.95 & 1.0 & 0.9719 & 735.01 \\
\bottomrule
\end{tabular}

\end{table}

\paragraph{Per-class F1 decomposition.}
Among the evaluation tasks, only empathy shows a gap of at least 10 points between Jev's accuracy and macro-F1, the signature of class-imbalanced predictions. Table~\ref{tab:b9} decomposes it: both Jev and Gemini 3.8 Flash collapse onto the no-exploration class, so the empathy failure is shared across model classes rather than specific to the decision model.

\begin{table}[!htbp]
\centering
\caption{Per-class precision, recall, and F1 on the empathy task, the one evaluation task with an accuracy-to-macro-F1 gap of at least 10 points.}
\label{tab:b9}
\small
\begin{tabular}{lrrrrrr}
\toprule
Task & Model & Class & Support & P & R & F1 \\
\midrule
Empathy & Jev & A & 166 & 0.444 & 0.024 & 0.046 \\
Empathy & Jev & B & 166 & 0.418 & 0.169 & 0.240 \\
Empathy & Jev & C & 166 & 0.363 & 0.922 & 0.520 \\
Empathy & Gem 3.8F & A & 166 & 0.389 & 0.042 & 0.076 \\
Empathy & Gem 3.8F & B & 166 & 0.417 & 0.151 & 0.221 \\
Empathy & Gem 3.8F & C & 166 & 0.360 & 0.910 & 0.515 \\
\bottomrule
\end{tabular}

\end{table}

\paragraph{Selection-aware bootstrap for the best-LLM comparison.}
The main-text comparator for $\Delta$F1 is the per-task best LLM, selected on the same test items the interval is computed on. Table~\ref{tab:b12} recomputes every interval with the best LLM re-selected inside each of the 10{,}000 paired resamples, so the interval carries the uncertainty of the selection step. One verdict changes: the character-tropes interval moves from spanning zero to $[-15.5, -1.4]$ points; the ideological-books ($[-6.5, +0.1]$) and dialect ($[-4.3, +9.8]$) intervals continue to span zero, and every already-significant deficit remains so.

\begin{table}[!htbp]
\centering
\caption{Per-task $\Delta$F1 (Jev minus best LLM, points) with the fixed-best interval and the selection-aware interval that re-selects the best LLM inside each resample.}
\label{tab:b12}
\small
\begin{tabular}{lrcc}
\toprule
Task & $\Delta$F1 & Fixed-best 95\% CI & Selection-aware 95\% CI \\
\midrule
Dialect & +4.2 & [-3.9, +11.5] & [-4.3, +9.8] \\
Discourse$^{d}$ & -11.6 & [-15.6, -7.5] & [-15.8, -8.4] \\
Emotion & -13.5 & [-17.4, -9.9] & [-17.3, -9.8] \\
Empathy & -4.5 & [-7.7, -1.3] & [-8.3, -2.9] \\
Figurative & -6.2 & [-9.0, -3.5] & [-9.3, -3.7] \\
Humor & -13.0 & [-17.9, -8.2] & [-18.2, -8.8] \\
Ideology (doc.) & -17.2 & [-21.1, -13.4] & [-21.1, -13.4] \\
Ideology (utt.) & -2.1 & [-5.9, +1.8] & [-6.5, +0.1] \\
Impl. Hate$^{d}$ & -15.7 & [-20.3, -11.3] & [-20.2, -11.5] \\
Misinfo & -4.3 & [-7.7, -0.9] & [-7.9, -2.5] \\
Persuasion (conv.) & -18.9 & [-23.6, -14.3] & [-23.7, -14.5] \\
Persuasion (utt.) & -11.6 & [-16.3, -7.0] & [-16.3, -7.1] \\
Politeness & -10.5 & [-15.2, -5.7] & [-15.7, -7.1] \\
Power & -12.9 & [-17.3, -8.7] & [-17.4, -9.6] \\
Sem. Change & -16.6 & [-22.0, -11.3] & [-22.5, -12.3] \\
Stance$^{d}$ & -2.4 & [-6.0, +1.2] & [-6.2, +0.6] \\
Toxicity & -12.8 & [-17.1, -8.5] & [-17.9, -9.7] \\
Tropes & -7.1 & [-14.4, +0.3] & [-15.5, -1.4] \\
\bottomrule
\end{tabular}

\end{table}

\paragraph{Bootstrap uncertainty for calibration and routing.}
The calibration and routing quantities in Section~4 are point estimates; Table~\ref{tab:b13} adds item-level bootstrap intervals (2{,}000 resamples) for Jev's per-task expected calibration error, coverage, and accuracy at the 0.9 threshold. Task-level bootstrap intervals for the across-task medians are reported in Section~4: Jev's median ECE interval $[0.108, 0.233]$ overlaps Opus 5's $[0.046, 0.137]$, and Jev's median Acc@0.9 interval $[0.691, 0.897]$ includes the 0.85 target, so the two headline calibration comparisons are directionally consistent but not decisively separated at the median level; the per-task empathy interval (ECE $[0.499, 0.585]$) is far from every other task's.

\begin{table}[!htbp]
\centering
\caption{Item-level bootstrap intervals for Jev's per-task calibration and routing quantities (2{,}000 resamples; -- marks an empty 0.9-confidence subset).}
\label{tab:b13}
\small
\begin{tabular}{lrccc}
\toprule
Task & $n$ & ECE [95\% CI] & Cov@0.9 [95\% CI] & Acc@0.9 [95\% CI] \\
\midrule
Dialect & 266 & 0.063 [0.060, 0.131] & 0.289 [0.237, 0.346] & 0.922 [0.855, 0.975] \\
Discourse$^{d}$ & 497 & 0.096 [0.077, 0.144] & 0.241 [0.203, 0.280] & 0.833 [0.766, 0.898] \\
Emotion & 498 & 0.281 [0.246, 0.324] & 0.450 [0.406, 0.494] & 0.705 [0.647, 0.763] \\
Empathy & 498 & 0.538 [0.499, 0.585] & 0.781 [0.745, 0.817] & 0.383 [0.334, 0.433] \\
Figurative & 500 & 0.039 [0.036, 0.074] & 0.564 [0.522, 0.606] & 0.975 [0.956, 0.993] \\
Humor & 500 & 0.174 [0.146, 0.227] & 0.202 [0.166, 0.238] & 0.782 [0.695, 0.857] \\
Ideology (doc.) & 498 & 0.125 [0.103, 0.174] & 0.335 [0.295, 0.380] & 0.856 [0.801, 0.909] \\
Ideology (utt.) & 498 & 0.124 [0.104, 0.173] & 0.376 [0.331, 0.418] & 0.824 [0.768, 0.874] \\
Impl. Hate$^{d}$ & 498 & 0.210 [0.178, 0.256] & 0.181 [0.149, 0.215] & 0.744 [0.652, 0.831] \\
Misinfo & 500 & 0.108 [0.086, 0.148] & 0.526 [0.482, 0.572] & 0.894 [0.856, 0.929] \\
Persuasion (conv.) & 434 & 0.185 [0.161, 0.243] & 0.187 [0.152, 0.224] & 0.691 [0.593, 0.794] \\
Persuasion (utt.) & 399 & 0.157 [0.126, 0.206] & 0.441 [0.393, 0.489] & 0.807 [0.750, 0.863] \\
Politeness & 498 & 0.101 [0.079, 0.147] & 0.195 [0.161, 0.229] & 0.897 [0.835, 0.954] \\
Power & 500 & 0.233 [0.208, 0.287] & 0.404 [0.358, 0.448] & 0.644 [0.576, 0.709] \\
Sem. Change & 344 & 0.170 [0.139, 0.227] & 0.224 [0.180, 0.273] & 0.909 [0.840, 0.971] \\
Stance$^{d}$ & 435 & 0.138 [0.116, 0.186] & 0.448 [0.402, 0.494] & 0.856 [0.807, 0.903] \\
Toxicity & 500 & 0.302 [0.266, 0.350] & 0.642 [0.600, 0.684] & 0.611 [0.559, 0.668] \\
Tropes & 114 & 0.126 [0.098, 0.207] & 0.000 [0.000, 0.000] & -- \\
\bottomrule
\end{tabular}

\end{table}

\paragraph{Confidence and human annotator disagreement.}
Politeness is the one grid task whose upstream release ships raw per-annotator scores (five annotators per item, 1--25 scale); all 498 test items match by request text. No model's stated confidence tracks human disagreement: the Spearman correlation between per-item confidence and the five-score standard deviation is at most $|\rho| = 0.14$ across the six models tested, and the only significant coefficients are positive (Table~\ref{tab:b14}). Accuracy is equally unrelated to disagreement (Jev $\rho = -0.01$, Opus 5 $\rho = -0.03$). On this task, annotator disagreement is an axis of item difficulty that neither the decision model's distribution nor verbalized confidence represents, the direct form of the gold-standard scoping in Section~3.3.

\begin{table}[!htbp]
\centering
\caption{Spearman correlation between per-item stated confidence and the standard deviation of the five human annotator scores on the politeness task.}
\label{tab:b14}
\small
\begin{tabular}{lrrr}
\toprule
Model & $n$ & Spearman $\rho$ & $p$ \\
\midrule
Jev & 498 & +0.067 & 0.137 \\
Gem 3.8F & 498 & +0.031 & 0.488 \\
Opus 5 & 495 & +0.144 & 0.001 \\
Fable 5.1 & 487 & +0.086 & 0.059 \\
Sonnet 5 & 498 & +0.095 & 0.034 \\
Gemma 31B & 498 & +0.073 & 0.103 \\
\bottomrule
\end{tabular}

\end{table}

\paragraph{Structured output as an interface control.}
The zero-invalid-answer property of the decision models could reflect the output interface rather than the model class. As a control, Gemini 3.8 Flash is rerun on all 18 tasks with its answer forced through a JSON schema whose answer field enumerates the task's options, with the same prompts and configuration otherwise, paired on identical items. Table~\ref{tab:b15} compares the two interfaces. Constrained output changes little: median evaluation-task macro-F1 moves from 0.669 to 0.676, with per-task changes from $-2.0$ to $+11.4$ points, the largest on the 114-option tropes task ($n = 113$), so the free-text protocol of the main runs does not meaningfully understate the LLM side. The interface guarantee itself transfers only partly: schema-constrained serving still returned 16 invalid answers on the paired item set (against 34 under free text and zero across all main-evaluation decision-model calls), because some provider responses fall back to unconstrained text.

\begin{table}[!htbp]
\centering
\caption{Interface control: Gemini 3.8 Flash under the main free-text protocol against a JSON-schema-constrained rerun on identical items (macro-F1 and accuracy $\times 100$; Invalid counts unparseable or schema-violating answers).}
\label{tab:b15}
\small
\setlength{\tabcolsep}{4pt}
\begin{tabular}{lrrrrrr}
\toprule
Task & $n$ & \multicolumn{2}{c}{Macro-F1} & \multicolumn{2}{c}{Accuracy} & Invalid \\
 & & free text & structured & free text & structured & free/struct.\ \\
\midrule
Dialect & 266 & 59.8 & 59.0 & 46.2 & 45.5 & 2/2 \\
Discourse$^{d}$ & 497 & 71.1 & 71.5 & 71.2 & 71.4 & 0/0 \\
Emotion & 498 & 56.5 & 60.3 & 57.2 & 60.4 & 0/0 \\
Empathy & 498 & 27.1 & 29.2 & 36.7 & 38.0 & 0/0 \\
Figurative & 500 & 92.5 & 93.4 & 92.6 & 93.4 & 0/1 \\
Humor & 500 & 67.7 & 67.6 & 67.8 & 68.0 & 0/1 \\
Ideology (doc.) & 498 & 82.6 & 83.8 & 82.5 & 83.7 & 0/0 \\
Ideology (utt.) & 498 & 63.2 & 63.3 & 63.3 & 63.3 & 0/1 \\
Impl. Hate$^{d}$ & 498 & 57.1 & 58.8 & 56.0 & 59.2 & 24/1 \\
Misinfo & 500 & 82.1 & 82.3 & 82.6 & 82.6 & 0/2 \\
Persuasion (conv.) & 434 & 72.5 & 72.9 & 73.5 & 73.7 & 0/2 \\
Persuasion (utt.) & 399 & 72.4 & 74.0 & 72.9 & 74.7 & 0/1 \\
Politeness & 498 & 66.9 & 66.7 & 67.1 & 66.7 & 0/2 \\
Power & 500 & 66.3 & 68.3 & 68.4 & 70.2 & 0/0 \\
Sem. Change & 344 & 84.9 & 85.6 & 84.9 & 85.5 & 0/1 \\
Stance$^{d}$ & 435 & 75.8 & 75.4 & 76.3 & 75.9 & 0/1 \\
Toxicity & 500 & 61.7 & 59.6 & 61.8 & 60.0 & 0/0 \\
Tropes & 113 & 24.3 & 35.8 & 31.0 & 41.6 & 8/1 \\
\bottomrule
\end{tabular}

\end{table}

\paragraph{The open decision-model ecosystem.}
Eleven open decision systems, tracked by a concurrent community leaderboard \citep{jevbench2026}, are run on the 17-task local grid under each system's own released inference code. Ten were released in the week after Jev's launch: Laya, a 0.4B ModernBERT-large encoder answering in one forward pass \citep{kishor2026laya}; Von, a 0.4B ModernBERT-large encoder trained with cross-entropy plus a Brier term \citep{panisa2026von}; Verdict, a 0.15B ModernBERT encoder trained the same way, with an explicit abstention option \citep{kumar2026verdict}; decider-0.8b and decider-2b, decoders trained for typed decisions on Qwen3.5 bases \citep{mapika2026decider}; SemIf, a frozen Qwen3.5-4B with a logit readout over option letters and no training \citep{lee2026semif}; Bespoke-Nimble-9B, a LoRA adapter on Qwen3.5-9B trained on roughly 3{,}000 contrastive examples \citep{bespoke2026nimble}; and Kev-0.8B, 4B, and 9B, LoRA adapters with a pointer head trained with cross-entropy on identical data at three base sizes \citep{palmer2026kev}. The eleventh, OpenDecision, is a decision interface released in the same week around a zero-shot natural-language-inference classifier from December 2024 \citep{wadhwa2026opendecision,laurer2023building}, so it measures what the interface alone delivers on a model that predates Jev. The two systems that refuse over-length prompts rather than truncate receive states tail-truncated to fit, SemIf at the same 1{,}536-token budget as the two local models of the main evaluation; SemIf's 16 answer slots also exclude the 23-option dialect task. Verdict abstains on 38.8\% of evaluation-task items; we score it, like every other system, as a forced choice over the declared options, renormalizing after dropping the abstention option. These runs were added after the analysis plan was filed and enter none of the main-text tests. Table~\ref{tab:b11} reports per-task macro-F1 with median macro-F1, ECE, and Brier over the evaluation tasks each system covers, and Figures~\ref{fig:accuracy} and~\ref{fig:calibration} place the open systems among the main-text models. Three regularities emerge. First, the commercial model keeps its lead, and its largest margin is the dialect task (64.0 against 46.0 for the best open system), but the gap narrows with base-model scale more than with decision training. The untrained 4B readout posts the highest open median macro-F1 (55.0), Kev rises from 41.9 to 50.8 to 53.3 as only the base grows from 0.8B to 9B, and the December 2024 classifier (42.7) exceeds all three encoders trained for decisions since (35.1 to 41.6), echoing at larger scale the main-text result that the 0.6B RLCD model beats its untrained base (36.1 against 27.8). Second, several open systems return better-calibrated probabilities than Jev, but the confidence each system reports is not always that probability. On the top-class probability, decider-0.8b and Von reach a median ECE of 0.081 and Verdict 0.086, against 0.178 for Jev on the same 14 tasks, the three best values among the decision systems, at median macro-F1 12 to 24 points below Jev's. The training recipe alone does not predict this, since decider-2b, trained with the same recipe as decider-0.8b, reaches 0.134. Von, however, reports the gap between its two highest probabilities as its confidence, and on that field its median ECE is 0.301, the worst in the class; Kev reports a confidence of its own whose median ECE (0.153 to 0.173) brackets Jev's 0.163. A researcher thresholding on a decision system's confidence should first check which quantity it is. Third, the empathy failure of Section~4 is a property of the construct, not the vendor: every open system from the 0.15B encoder to the 9B adapters lands between 19.8 and 35.7 macro-F1 on empathy, and SemIf and Nimble state mean confidence above 0.94 on that task. All eleven systems annotated their grids without an API account, six of them on a laptop and the other five on a single A100 GPU.
\begin{table}[!htbp]
\centering
\caption{Per-task macro-F1 ($\times 100$) for the commercial decision model (Jev), the two local models of the main evaluation, and the eleven open decision systems (marked $^{\dagger}$, ordered by size), each under its own released inference code. Medians are over the evaluation tasks each system covers ($^{d}$ marks pilot tasks, excluded from medians; -- marks the dialect cell outside SemIf's 16 answer slots; tropes exceeds every letter-readout architecture and is omitted). ECE is computed on each system's reported confidence and, in the $p_{\mathrm{top}}$ row, on its top-class probability.}
\label{tab:b11}
\footnotesize
\setlength{\tabcolsep}{3pt}
\begin{tabular}{lrrrrrrrrrrrrrr}
\toprule
Task & \rot{Jev} & \rot{Qwen3-0.6B-RLCD} & \rot{Qwen3-0.6B base} & \rot{Verdict-0.15B$^{\dagger}$} & \rot{Laya-0.4B$^{\dagger}$} & \rot{NLI-0.4B$^{\dagger}$} & \rot{Von-0.4B$^{\dagger}$} & \rot{decider-0.8b$^{\dagger}$} & \rot{Kev-0.8B$^{\dagger}$} & \rot{decider-2b$^{\dagger}$} & \rot{SemIf-4B$^{\dagger}$} & \rot{Kev-4B$^{\dagger}$} & \rot{Nimble-9B$^{\dagger}$} & \rot{Kev-9B$^{\dagger}$} \\
\midrule
Dialect & 64.0 & 5.7 & 2.7 & 0.2 & 4.6 & 23.4 & 8.6 & 26.5 & 26.4 & 31.3 & -- & 33.6 & 46.0 & 38.5 \\
Emotion & 48.4 & 49.7 & 28.1 & 47.2 & 48.2 & 65.9 & 76.6 & 69.3 & 53.4 & 81.0 & 49.0 & 54.1 & 52.3 & 51.8 \\
Figurative & 86.3 & 30.1 & 27.4 & 15.1 & 20.5 & 11.5 & 39.3 & 37.8 & 35.8 & 40.2 & 77.3 & 65.2 & 70.0 & 71.8 \\
Humor & 56.3 & 44.2 & 33.1 & 34.7 & 52.7 & 42.0 & 33.5 & 58.6 & 53.5 & 55.4 & 51.4 & 63.0 & 62.4 & 59.6 \\
Ideology (utt.) & 63.3 & 19.6 & 19.7 & 33.0 & 40.6 & 29.9 & 34.2 & 45.0 & 31.2 & 35.7 & 56.0 & 40.9 & 56.3 & 50.8 \\
Impl. Hate$^{d}$ & 43.9 & 28.0 & 17.6 & 27.3 & 22.9 & 22.2 & 14.1 & 23.3 & 27.8 & 33.8 & 44.3 & 38.5 & 44.8 & 35.9 \\
Misinfo & 79.6 & 60.2 & 34.2 & 51.4 & 64.6 & 71.3 & 62.5 & 64.8 & 54.6 & 71.6 & 76.2 & 65.9 & 68.1 & 63.6 \\
Persuasion (utt.) & 60.7 & 39.2 & 6.3 & 22.4 & 30.0 & 43.5 & 56.2 & 50.2 & 35.0 & 48.3 & 55.0 & 49.1 & 58.9 & 54.7 \\
Sem. Change & 68.3 & 44.9 & 41.3 & 45.6 & 42.6 & 44.6 & 52.1 & 57.6 & 49.4 & 48.3 & 55.2 & 41.7 & 47.5 & 41.1 \\
Stance$^{d}$ & 73.4 & 48.2 & 16.7 & 28.5 & 49.9 & 44.5 & 43.7 & 50.4 & 42.5 & 37.6 & 55.8 & 62.0 & 73.8 & 68.1 \\
Discourse$^{d}$ & 59.5 & 27.5 & 8.2 & 14.4 & 30.5 & 29.3 & 28.0 & 32.2 & 20.8 & 44.8 & 48.8 & 47.5 & 53.0 & 50.3 \\
Empathy & 26.9 & 30.7 & 16.9 & 25.5 & 35.7 & 27.2 & 23.5 & 35.4 & 32.0 & 32.4 & 19.8 & 20.7 & 26.2 & 22.4 \\
Persuasion (conv.) & 56.1 & 44.6 & 33.3 & 39.1 & 50.3 & 51.0 & 51.7 & 53.4 & 49.6 & 55.8 & 58.7 & 58.1 & 58.1 & 56.8 \\
Politeness & 57.3 & 27.0 & 22.6 & 27.5 & 30.8 & 43.4 & 40.4 & 50.2 & 46.1 & 42.9 & 57.3 & 52.4 & 53.6 & 45.3 \\
Power & 58.1 & 38.8 & 43.7 & 46.8 & 50.1 & 54.1 & 58.0 & 39.1 & 37.5 & 47.4 & 54.9 & 58.2 & 51.5 & 54.7 \\
Toxicity & 50.2 & 33.3 & 41.9 & 49.9 & 43.3 & 36.6 & 38.5 & 37.1 & 47.2 & 36.9 & 46.7 & 40.4 & 53.3 & 40.3 \\
Ideology (doc.) & 65.4 & 22.6 & 17.4 & 35.4 & 29.0 & 25.3 & 28.5 & 23.5 & 37.7 & 32.5 & 36.3 & 45.3 & 53.9 & 54.7 \\
\midrule
Median F1 & 59.4 & 36.1 & 27.8 & 35.1 & 41.6 & 42.7 & 39.9 & 47.6 & 41.9 & 45.2 & 55.0 & 50.8 & 53.8 & 53.3 \\
Median ECE & 0.163 & 0.128 & 0.144 & 0.086 & 0.275 & 0.204 & 0.301 & 0.081 & 0.153 & 0.134 & 0.217 & 0.171 & 0.248 & 0.173 \\
Median ECE ($p_{\mathrm{top}}$) & 0.178 & 0.128 & 0.144 & 0.086 & 0.193 & 0.204 & 0.081 & 0.081 & 0.176 & 0.134 & 0.217 & 0.140 & 0.248 & 0.109 \\
Median Brier & 0.534 & 0.662 & 0.688 & 0.668 & 0.693 & 0.688 & 0.582 & 0.619 & 0.628 & 0.637 & 0.572 & 0.616 & 0.663 & 0.574 \\
\bottomrule
\end{tabular}

\end{table}

\paragraph{Sensitivity to the empathy task, and cost at accuracy parity.}
Removing the empathy task, the grid's largest calibration failure, lowers Jev's median expected calibration error from 0.157 to 0.142 and raises median accuracy in the 0.9-confidence subset from 0.815 to 0.824; the calibration and routing conclusions are unchanged. On the three evaluation tasks where the accuracy difference to the per-task best LLM spans zero, Jev's measured cost advantage is 35$\times$ (ideological books, \$0.016 against \$0.568 per 1{,}000 items), 44$\times$ (Indian English dialect, \$0.048 against \$2.135), and 141$\times$ (character tropes, \$0.222 against \$31.40). The full-grid results remain the headline; ideological books and dialect are where the cost advantage comes with no accuracy deficit, while the character-tropes interval falls below zero once the best LLM is re-selected inside each resample (Table~\ref{tab:b12}).

\end{document}